\documentclass{article}

\usepackage{microtype}
\usepackage{graphicx}
\usepackage{subcaption}
\usepackage{booktabs} 

\usepackage{hyperref}

\usepackage[accepted]{icml2026}

\usepackage{amsmath}
\usepackage{amssymb}
\usepackage{mathtools}
\usepackage{amsthm}

\usepackage[capitalize,noabbrev]{cleveref}

\usepackage{multirow}
\usepackage{enumitem}
\usepackage{xcolor}
\renewcommand{\algorithmiccomment}[1]{\textcolor{gray}{\# #1}}
\usepackage{listings}

\theoremstyle{plain}

\theoremstyle{definition}

\theoremstyle{remark}

\icmltitlerunning{Efficient Test-time Inference for Generative Planning Models}

\begin{document}

\twocolumn[
  \icmltitle{Efficient Test-time Inference for\\ Generative Planning Models with OCL Search}




  \begin{icmlauthorlist}
    \icmlauthor{Robert Gieselmann}{amazon}
    \icmlauthor{Mihai Samson}{amazon}
    \icmlauthor{Federico Pecora}{prevamazon}
    \icmlauthor{Jeremy L. Wyatt}{amazon}
  \end{icmlauthorlist}

  \icmlaffiliation{amazon}{Amazon}
  \icmlaffiliation{prevamazon}{Work done while at Amazon}

  \icmlcorrespondingauthor{Robert Gieselmann}{robgie@amazon.com}

  \icmlkeywords{Machine Learning, ICML}

  \vskip 0.3in
]



\printAffiliationsAndNotice{}  

\begin{abstract}
Generative models have emerged as a powerful paradigm for AI planning, yet their performance remains constrained by the training data distribution. One approach is to improve generated solutions during inference by scaling test-time compute. A more efficient alternative is to optimize the inference process itself. In this paper, we show that a modified version of a classical Open-Closed List (OCL) search provides just such an efficient inference procedure. Our algorithm synergizes two learned components: a generative model that performs fast rollouts from intermediate states and a heuristic model that prioritizes among candidate reasoning paths. Key contributions include novel exploration control mechanisms and integration of learned models within the OCL framework. Across multiple combinatorial planning domains, our approach outperforms both neurosymbolic search baselines and classical solvers in computational efficiency and solution quality.
\end{abstract}

\section{Introduction}\label{sec:introduction}

Automated planning seeks action sequences that transform an initial state into one satisfying goal conditions. Deep generative models have emerged as a promising paradigm for plan generation, offering fast synthesis across diverse domains~\cite{rossetti2024learning}. However, solution quality is bounded by the training data, and collecting large-scale optimal datasets is often infeasible; even seemingly simple domains like Blocksworld are NP-hard to solve optimally~\cite{SLANEY2001119}.

An alternative is to spend additional compute at inference time. Best-of-N sampling~\cite{stiennon2020learning} repeatedly queries a generative model and returns the best generation, but does not balance exploration and exploitation to find better solutions at lower cost. Search algorithms like Monte Carlo Tree Search (MCTS) have become popular for test-time compute in formal reasoning domains such as theorem proving~\cite{chen2024alphamath, Hubert2025Olympiad} or code generation~\cite{antoniades2025swesearch}. However, MCTS methods typically employ UCT-style selection (e.g., PUCT; \citealp{rosin11}) with exploration incentives growing with parent visitation counts. Because every traversal originates at the root, deeper nodes are visited less frequently, often yielding wide but shallow trees. This depth imbalance is problematic for long-horizon planning, where critical decisions deep in the tree remain underexplored.

Open-Closed List (OCL) algorithms \citep{hart68, Valenzano_Xie_2016} like A* select nodes globally and underpin state-of-the-art symbolic planners \citep{HELMERT2003219}. However, unlike MCTS where generative models can serve as rollout policies, standard OCL provides no mechanism for fast candidate generation nor does it compensate for inadmissible heuristics. When trained on suboptimal demonstrations, learned heuristics overestimate cost-to-go, disproportionately inflating estimates for states far from the goal and causing greedy behavior that favors near-goal nodes. Moreover, querying generative models is computationally expensive, necessitating selective invocation.

We present \textsc{OCLGen}, a compute-efficient test-time search algorithm for autoregressive planning models that substantially improves plan quality. Our approach addresses these challenges through three innovations: (1)~\emph{depth-partitioned selection} maintains separate open lists per depth level, ensuring balanced exploration despite overestimating heuristics; (2)~\emph{truncated rollouts with adaptive expansion} uses the generative model for fast multi-step synthesis while branching at low-confidence decision points; and (3)~\emph{distributional heuristic estimation} ranks nodes by a lower-percentile cost-to-go estimate, targeting their best attainable outcome.

Experiments on four classical planning domains demonstrate that \textsc{OCLGen} converges to shorter plans significantly faster than baselines. On problems with known optimal solutions, our method achieves 87.3\% optimality across domains, compared to 49.8\% for MCTS given the same compute budget. \textsc{OCLGen} also provides an effective foundation for iterative self-improvement: after three refinement iterations, our method achieves 100\% optimal plans on Blocksworld and 94.7\% on Sokoban, compared to 13.5\% and 51.3\%, respectively, for the base model with Best-of-N sampling given the same runtime budget. In summary, our contributions include:
\begin{itemize}[itemsep=0.1pt, topsep=0pt, partopsep=0pt]
    \item A modernized OCL framework with autoregressive models for fast rollout generation.
    \item A distributional approach for deploying overestimating heuristics in search.
    \item Comprehensive evaluation across four domains with ablations validating each design choice.
    \item Demonstration of \textsc{OCLGen}'s effectiveness within a recursive self-improvement framework.
\end{itemize}
\section{Related Work}\label{sec:background}

\paragraph{Generative Models for AI Planning}

Deep generative models have emerged as a promising approach for planning by learning to generate solutions from collections of solved instances. Recent work has explored both pre-trained large language models (LLMs) and smaller domain-specialized transformers for plan generation. While LLMs can be fine-tuned for planning tasks~\citep{pallagani2023plansformer}, concerns about inference costs and reliability have motivated training smaller transformers from scratch on planning data. These approaches typically formulate planning as autoregressive sequence generation, predicting plans token by token. For example, PlanGPT~\citep{rossetti2024learning} trains a GPT2 model on plans from a domain-independent planner by splitting individual actions into sequences of operator name and object tokens. Extensions include integrating action validators during generation~\citep{rossetti2024enhancing}, repairing invalid plans via local search~\citep{tummolo2024integrating}, and leveraging symmetries for improved generalization~\citep{fritzsche2025symmetry}. Our work adopts the model architecture and tokenization scheme from~\citep{rossetti2024learning} and contributes an efficient inference algorithm.

\paragraph{Test-time Search} 

Test-time inference in generative models is increasingly framed as a search over candidate solutions or reasoning paths. Recent frameworks include Tree of Thoughts \citep{yao2023tree}, which applies breadth-first search or depth-first search to deliberative reasoning, and rStar \citep{guan2025rstar}, which leverages MCTS for multi-step reasoning. In formal domains, DeepSeek-Prover-V1.5 \citep{xin2025deepseekproverv} integrates MCTS with proof assistant feedback, while PG-TD \citep{zhang2023planning} combines AlphaZero-style search with test case execution for verified code generation. While most of these methods target general LLM reasoning, in this work we focus on smaller, domain-specialized planning models, where the same search principles can substantially improve output quality.

\paragraph{Self-Improvement via Search}

Test-time search can serve as a standalone inference procedure or underpin recursive self-improvement, where search-generated solutions provide supervision for retraining. Combining search and learning in a self-improvement loop dates back to early checkers programs~\citep{samuelscheckers}, and has since been applied to classical planning via heuristic search with learned neural heuristics~\citep{JABBARIARFAEE20112075, groshev2018learning}. The idea gained widespread attention with AlphaGo~\citep{silver2016mastering} and AlphaZero~\citep{silver2017mastering}, which coupled MCTS with deep networks trained via self-play. Similarly, the Expert Iteration (ExIt) framework~\citep{anthony2017thinking} formalizes this loop, where an \emph{expert} search iteratively supervises an \emph{apprentice} policy. Analogous frameworks have recently been applied to fine-tuning transformers for reasoning~\citep{lehnert2024beyond, zhang2024rest, chen2024alphamath}, theorem proving~\citep{xin2025deepseekproverv}, and multi-step reasoning~\citep{guan2025rstar}. The generative planning framework in \citet{gieselmann2026selfimprovementfasthighqualityplan} combines Best-of-N sampling with graph processing to refine a transformer in a recursive loop that iterates between data curation and model finetuning. We build on this work, focusing on efficient test-time search which further accelerates self-improvement.

\section{Preliminaries}

\subsection{Classical Planning in PDDL}\label{sec:ai_planning}

We focus on classical single-agent planning in deterministic, fully observable, discrete environments. A planning problem is specified by a finite set of objects $\mathcal{O}$ and a finite set of predicates $\mathcal{P}$, where each predicate $p \in \mathcal{P}$ has an associated arity $\mathrm{arity}(p)$. Grounded atoms are formed by instantiating predicates with objects:
$\mathcal{F} = \bigcup_{p \in \mathcal{P}} \left\{\, p(o_1, \ldots, o_{\mathrm{arity}(p)}) \mid o_1, \ldots, o_{\mathrm{arity}(p)} \in \mathcal{O} \,\right\}.$
A state $s \in \mathcal{S} = 2^\mathcal{F}$ represents the set of atoms that hold true in a given world configuration. Actions $a \in \mathcal{A}$ are grounded operators $a = \langle \mathrm{pre}(a), \mathrm{add}(a), \mathrm{del}(a) \rangle$, where preconditions $\mathrm{pre}(a) \subseteq \mathcal{F}$, add effects $\mathrm{add}(a) \subseteq \mathcal{F}$, and delete effects $\mathrm{del}(a) \subseteq \mathcal{F}$ determine applicability and outcome. An action $a$ is applicable in state $s$ when $\mathrm{pre}(a) \subseteq s$, yielding successor state $F(s,a) = (s \setminus \mathrm{del}(a)) \cup \mathrm{add}(a)$. Goals $s_g \in \mathcal{G} \subseteq 2^\mathcal{F}$ are partial state descriptions. A state $s$ satisfies goal $s_g$ if $s_g \subseteq s$. A plan $\tau = (a_0, \ldots, a_{T-1})$ is valid if each action $a_t$ is applicable in $s_t$, and executing the sequence from $s_0$ via $s_{t+1} = F(s_t, a_t)$ reaches a final state $s_T$ satisfying the goal. In this work, we define optimality as minimum plan length.

The Planning Domain Definition Language (PDDL)~\citep{mcdermott1998pddl,fox2003pddl21} serves as the standard formalism in AI planning, underpinning benchmarks such as the International Planning Competition~\citep{vallati2015ipc}. A PDDL domain defines predicates, typed objects, and operators specified through preconditions and effects. Individual problem instances declare concrete objects, an initial state, and goal conditions. We use PDDL as the formal language to formulate problems, states, and plans while leveraging its structure to tokenize planning problems for autoregressive sequence generation.

\subsection{Autoregressive Planning Models}\label{sec:generative_model}

We investigate test-time inference for generative models $\pi_\theta$ that produce distributions over actions conditioned on an initial state $s_0 \in \mathcal{S}$, a goal specification $s_g \in \mathcal{G}$, and the sequence of previous actions $a_{<t} \in \mathcal{A}^{t}$, i.e., $\pi_\theta(a_t \mid s_0, s_g, a_{<t})$. Following \citet{rossetti2024learning}, we model $\pi_\theta$ as a decoder-only transformer~\cite{radford2019language} that represents states and actions by tokenizing PDDL descriptions of predicates, operator names and objects. This representation enables the model to navigate the combinatorially large space of grounded actions using a compact, fixed-size vocabulary. Actions are split into tokens and generated autoregressively.

To illustrate, consider the Blocksworld domain --- a classical planning benchmark where the objective is to rearrange a set of blocks (e.g., \texttt{b1}, \texttt{b2}, \texttt{b3}) into a specified goal configuration using four operators: \texttt{pickup} and \texttt{putdown} for moving blocks to and from the table, and \texttt{stack} and \texttt{unstack} for placing or removing blocks on top of one another. In this domain, the grounded action \texttt{unstack b1 b2} (removing block \texttt{b1} from atop block \texttt{b2}) is encoded as a sequence comprising the operator token \texttt{unstack} followed by argument tokens \texttt{b1} and \texttt{b2}. At each generation step, the newly predicted token is concatenated to the input sequence for the subsequent forward pass. The vocabulary is derived directly from PDDL domain and problem files, comprising predicate names, object identifiers, type declarations, and special delimiter tokens. A domain-specific upper bound on the number of objects ensures bounded representations.

\section{Efficient Test-time Search for Autoregressive Planning}\label{sec:method}

Using the generative model $\pi_\theta$ defined in Section~\ref{sec:generative_model}, we develop an efficient test-time search algorithm. Our objective is to minimize plan length while generalizing across a distribution of problem instances $P_{\mathcal{S} \times \mathcal{G}}$. For each planning domain, a model $\pi_\theta$ is trained via supervised learning on $\mathcal{D}_\text{train}$, a dataset of suboptimal plans generated by running a symbolic solver on a set of instances drawn from $P_{\mathcal{S} \times \mathcal{G}}$.

\subsection{OCL Search with Generative Models}\label{sec:algorithm_overview}

We introduce \textsc{OCLGen}, an efficient test-time search algorithm for generative planning models built on the Open-Closed List (OCL) framework \citep{hart68, Valenzano_Xie_2016}. This framework represents a general class of graph search algorithms that includes, e.g., $A^*$. OCL search maintains an open list of frontier nodes pending exploration and a closed list that tracks already-visited states to prevent redundant expansions. At each iteration, a node $n$ is selected from the open list $\mathcal{O}$ according to a priority function $f(n) = g(n) + h(n)$, where $g(n)$ denotes the incurred path cost and $h(n)$ the heuristic cost-to-go. The selected node is then moved to the closed list $\mathcal{C}$ and expanded to generate successors. Crucially, the framework detects when a node is revisited via a shorter path from the root, updates its $g(n)$ value accordingly, and moves it back to $\mathcal{O}$.

\paragraph{Modifications.}
To adapt OCL search for generative planning models, we introduce several modifications to the standard framework. 
First, we introduce \emph{depth-partitioned selection}, maintaining separate open lists per depth level (determined by~$g(n)$) to balance exploration across the solution depth and counteract systematic heuristic overestimation. 
Second, we perform node expansions via \emph{truncated rollouts} using the generative model~$\pi_\theta$ rather than single-step transitions. This allows us to rapidly generate sequences of successor nodes, which effectively prunes large or combinatorial action spaces. 
Within each rollout, we further apply \emph{adaptive expansion}, using the generative model's token confidence to identify critical decision points. These branching points are expanded immediately, broadening exploration when the model is uncertain and narrowing it when the model is confident. 
Finally, we integrate a learned distributional heuristic model~$h_\phi$ to guide node selection. 

Altogether, these components modernize OCL graph search by combining even depth-level coverage with heuristic value guidance, while dynamically constraining breadth based on model confidence. We detail each modification below.

\subsection{Depth-Partitioned Selection}\label{sec:method:depth_selection}

Heuristic models optimized to predict the cost-to-go $h(n)$ from suboptimal training data systematically overestimate the true cost-to-go $h^*(n)$. To observe how this biases selection, we model the resulting bias as multiplicative, $h(n) \approx \alpha\, h^*(n)$ for some $\alpha > 1$. 

Consider a pathological case: two unvisited frontier nodes, $n_1$ and $n_2$, situated on different branches but projected to yield solutions of identical total length $L$ (i.e., $g(n_1) + h^*(n_1) = g(n_2) + h^*(n_2) = L$). Assuming uniform unit action costs, the occured cost $g(n)$ equals the node's depth $d$. Suppose $n_1$ is shallower than $n_2$ ($g(n_1) < g(n_2)$). Under our bias model, the priority function evaluates to:
\begin{equation}
  f(n) = g(n) + \alpha(L - g(n)) = \alpha L - (\alpha - 1)\, g(n) .
  \label{eq:f_bias}
\end{equation}
Because $f(n)$ is strictly \emph{decreasing} with respect to $g(n)$, the scores satisfy $f(n_2) < f(n_1)$. Consequently, selecting the node with the lowest $f$-score from a global open list $\mathcal{O}$ (as in standard $A^*$) strictly prefers the deeper node $n_2$, despite both offering paths to equally good solutions. The search thus systematically over-commits to deeper, arbitrary branches, leaving promising shallower alternative routes—where course corrections are most valuable—unexpanded.

\begin{figure}[t]
    \centering
    \includegraphics[width=\linewidth]{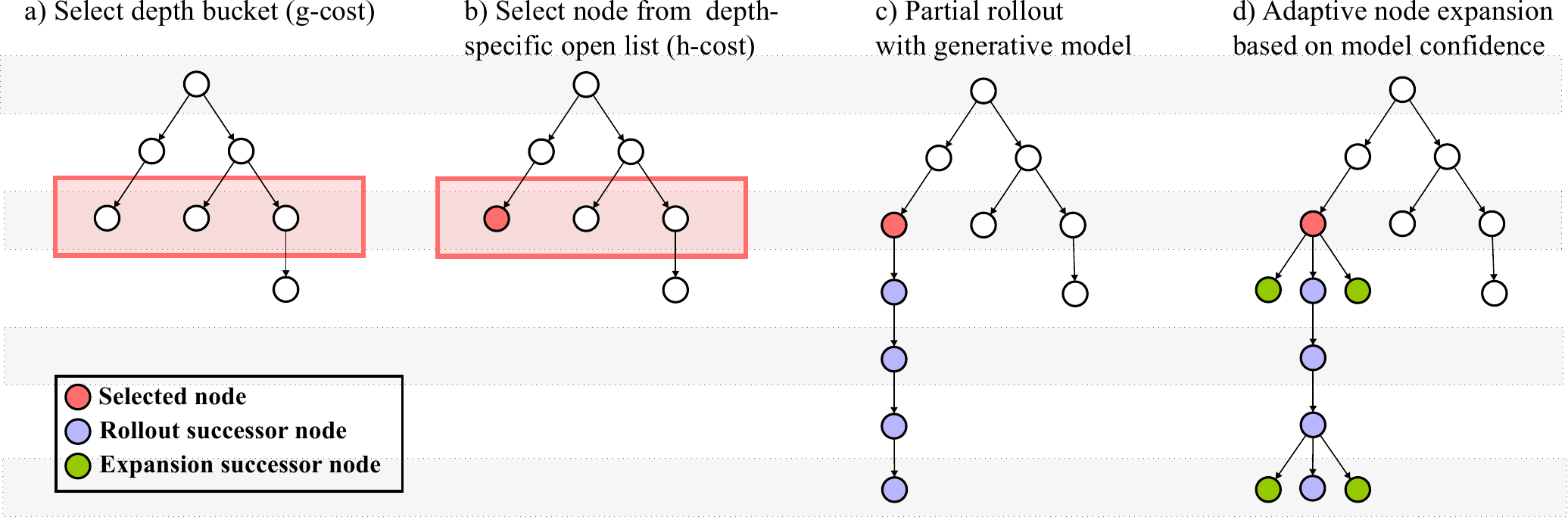}
    \caption{Illustration of one iteration with OCLGen. Note that we maintain a graph structure of states and transitions that is updated after every search iteration.}
    \label{fig:search_iteration}
\end{figure}

To mitigate this, we partition the unvisited nodes by depth, maintaining a separate open list $\mathcal{O}_d$ for each depth level $d$ (where $g(n) = d$). At each iteration, we first select a depth level $d_\text{s}$ (Figure~\ref{fig:search_iteration}a), then extract a node $n_\text{s}$ from $\mathcal{O}_{d_s}$ based on the heuristic estimate $h_\phi$ (Figure~\ref{fig:search_iteration}b). This two-stage selection ensures even exploration across potential solution lengths. Crucially, at a fixed depth $d$, pairwise comparisons within $\mathcal{O}_d$ depend only on the ranking signal $\alpha h^*(n)$. Thus, our learned heuristic $h_\phi$ provides a useful ranking signal within each depth level despite its absolute overestimation.

We investigate two strategies for selecting $d_\text{s}$ from $\{0, 1, \ldots, d_{\max}-1\}$, where $d_{\max}$ is the depth of the shallowest goal state found so far (or the maximum graph depth if no solution exists): \emph{Uniform selection} samples $d_\text{s}$ uniformly at random, while \emph{Scan selection} iterates sequentially from $d = 0$ to $d_{\max}-1$, ensuring systematic coverage across all levels.

\subsection{Confidence-based Node Expansion}\label{sec:method:expansion}
Having selected a depth level $d_\text{s}$ and extracted the highest-scoring node $n_s \in \mathcal{O}_{d_s}$ according to the heuristic $h_\phi$, we perform a truncated rollout starting from $n_\text{s}$ with the generative model $\pi_\theta$ (Figure~\ref{fig:search_iteration}c), producing a partial plan segment $(a_0, a_1, \ldots)$. The rollout is capped at a fixed number of tokens, chosen shorter than the longest plan's token sequence in the training data; this reduces computational overhead while still providing useful search guidance. We then decide, action by action, where to branch. For each generated action in the rollout, we compute a confidence value by taking the minimum probability assigned to any of the sampled tokens comprising that action (operator and argument tokens). We then apply \emph{adaptive expansion} (Figure~\ref{fig:search_iteration}d): for each node $n_i$ and outgoing action $a_i$ along the rollout where the action's confidence value falls below a threshold $\tau_\text{conf}$, we generate valid successor states as in classical node expansion and move node $n_i$ to $\mathcal{C}$. Nodes whose outgoing action has confidence above $\tau_\text{conf}$ are not expanded, concentrating computational resources on the critical decision points where exploration is most valuable. This contrasts with best-of-N sampling, which redraws complete plans and repeatedly regenerates the high-confidence segments shared across samples. Low confidence flags the decision points worth exploring; one plausible source is that the training data contains multiple alternative solutions to similar problems, leading the model to spread probability across competing actions.

To quantify the resulting savings, assume a constant branching factor $b$. Exhaustive expansion explores $\sum_{k=1}^{d} b^k$ nodes to reach depth $d$. Adaptive expansion instead follows confident actions greedily (local branching factor 1) and branches exhaustively otherwise (local branching factor $b$). If the model is confident at fraction $\gamma \in [0,1]$ of decision steps, the expected branching factor becomes
\begin{equation}
  \tilde{b} = (1 - \gamma)\,b + \gamma
  \label{eq:eff_branching}
\end{equation}
and, assuming branching decisions are approximately independent across nodes, the expected number of explored nodes to depth $d$ becomes $\sum_{k=1}^{d} \tilde{b}^{\,k}$. At $\gamma = 0$, Equation~\eqref{eq:eff_branching} recovers full expansion; at $\gamma = 1$, the sum collapses to $d$ (a single path). Because the node count grows as $\tilde{b}^{\,k}$, even a modest $\gamma$ compounds into substantial savings with depth, making rollouts with adaptive expansion especially effective in domains with large combinatorial action spaces.

All newly generated but unexpanded nodes are assigned to their respective depth-partitioned open lists $\mathcal{O}_d$ (or the closed list $\mathcal{C}$ if their state has already been visited). Consistent with standard OCL search, we update $g$-costs and reassign nodes to the appropriate depth list $\mathcal{O}_d$.

\subsection{Distributional Heuristic Model}\label{sec:method:heuristic}

We frame cost-to-go prediction as a discrete classification task over the set of possible remaining plan lengths $\mathcal{L} = \{0, 1, \ldots, L_{\max}\}$. The model, parameterized by $\phi$, outputs a conditional probability distribution $P_\phi(c \mid s, s_g)$ representing the predicted probability that the true remaining plan length from state $s$ to goal $s_g$ is exactly $c \in \mathcal{L}$. This model is trained via maximum likelihood estimation using a standard cross-entropy loss on $\mathcal{D}_\text{train}$. The scalar heuristic value $h_\phi(n)$ used downstream for search is extracted as a summary statistic from this distribution.

Because the model is trained on suboptimal demonstrations, $P_\phi$ concentrates on suboptimal plan lengths, leaving the true optimal cost-to-go $h^*$ in its lower tail. Depth partitioning (Section~\ref{sec:method:depth_selection}) already neutralizes the bias that overestimation induces when comparing nodes across depths; the choice of summary statistic plays a complementary role within a depth. Since search ultimately retains only the best plan reachable from a node, we want $h_\phi$ to rank nodes by their best attainable completion rather than their typical, suboptimal one. We therefore summarize $P_\phi$ by its lower $k$-th percentile rather than the mode, targeting the optimistic tail. While selecting the $k$-th percentile is not universally more optimistic than the mode for all distribution shapes, it empirically provides a useful ranking signal in our domains (see Appendix~\ref{app:cost_distributions}).

For unexpanded nodes $n_i$ along a rollout (where confidence exceeded $\tau_\text{conf}$), we minimize forward passes through the heuristic model by backing up the estimate from the nearest downstream expanded node $n_j$. Specifically, we set $h(n_i) = h_\phi(n_j) + g(n_j) - g(n_i)$, unless $n_i$ is the final state of the rollout.

\subsection{Base Model Improvements via Action Compilation}\label{sec:base_model_improvements}

Building on \citet{rossetti2024learning}, we introduce a new training data augmentation strategy that samples a random offset for each ground-truth plan, computes the corresponding intermediate state using a formal plan validator \citep{howey2004val}, and constructs training examples from these intermediate states paired with their remaining action sequences. This exposes the model to diverse starting states and shorter planning horizons, improving generalization necessary when calling the model on intermediate states within the search graph. We apply this strategy when training both policy and heuristic models. Empirical validation is provided in Appendix~\ref{app:action_compilation}.

\subsection{Algorithm Overview}\label{sec:method:overview}

\begin{algorithm}[tb]
\footnotesize
\caption{Depth-partitioned OCL for Generative Models}\label{alg:search}
    \textbf{Input}: Initial state $s_0$, goal $s_g$, iterations $n_\text{iters}$, 
    initial rollouts $N_\text{init}$, generative model $\pi_\theta$, heuristic model $h_\phi$
    \begin{algorithmic}
        \STATE \algorithmiccomment{Initialize empty search graph}
        \STATE $G \gets $\texttt{SearchGraph}()
        \STATE \algorithmiccomment{Add root node to graph}
        \STATE $G\texttt{.add}$(\texttt{Node}($s_0$, g = $0$, h = $h_\phi(s_0, s_g)$))
        \STATE \algorithmiccomment{Execute initial rollouts to populate graph}
        \STATE $G \gets \texttt{initial\_rollouts}$($G$, $\pi_\theta$, $s_0$, $s_g$, $N_\text{init}$)
        \STATE \algorithmiccomment{Main search loop}
        \FOR{$i_\text{iter}$ in $1 \ldots n_\text{iters}$}
            \STATE \algorithmiccomment{Select a depth level}
            \STATE $d_\text{s} \gets \texttt{select\_depth}$($G$, $i_\text{iter}$) \hfill \textcolor{blue}{(Fig.~\ref{fig:search_iteration}a)}
            \STATE \algorithmiccomment{Select node from depth-specific open list}
            \STATE $n_\text{s} \gets \texttt{select\_node}$($G$, $d_\text{s}$) \hfill \textcolor{blue}{(Fig.~\ref{fig:search_iteration}b)}
            \STATE \algorithmiccomment{Generate new nodes from rollout}
            \STATE $S_\text{roll} \gets \texttt{partial\_rollout}$($\pi_\theta$, $n_\text{s}$, $s_g$) \hfill \textcolor{blue}{(Fig.~\ref{fig:search_iteration}c)}
            \STATE \algorithmiccomment{Generate new child nodes by expansion}
            \STATE $S_\text{exp} \gets \texttt{adaptive\_expansion}$($n_\text{s}$, $s_g$, $S_\text{roll}$) \hfill \textcolor{blue}{(Fig.~\ref{fig:search_iteration}d)}
            \STATE \algorithmiccomment{Query $h_\phi$ on expanded nodes; assign heuristics via backpropagation to remaining rollout nodes}
            \STATE $\mathcal{H} \gets \texttt{assign\_heuristics}$($S_\text{exp}$, $S_\text{roll}$, $h_\phi$, $s_g$)
            \STATE \algorithmiccomment{Add new nodes to graph}
            \FOR{$(s_i, h_i)$ in $\mathcal{H}$}
                \STATE $G\texttt{.add}$(\texttt{Node}($s_i$, g = $g(\text{parent}(s_i)){+}1$, h = $h_i$))
            \ENDFOR
        \ENDFOR
    \STATE \textbf{return} $G$\texttt{.get\_best\_plan()}
  \end{algorithmic}
\end{algorithm}

We structure \textsc{OCLGen} as an anytime algorithm --- one that returns a valid solution at any point and continues refining it as more compute becomes available. Initially, we execute $N_\text{init}$ full rollouts from $s_0$ to establish baseline solutions and populate the search graph with goal-reaching trajectories. The algorithm then iteratively refines these solutions through depth-partitioned search until a compute budget is reached (e.g., a maximum number of iterations or wall-clock time). Algorithm~\ref{alg:search} provides pseudocode for the main search loop, and Figure~\ref{fig:search_iteration} illustrates each step visually. Each iteration proceeds as follows: (a) \texttt{select\_depth} chooses a depth level to focus expansion; (b) \texttt{select\_node} picks a promising node from the corresponding open list; (c) \texttt{partial\_rollout} generates a truncated trajectory from the selected node toward $s_g$; (d) \texttt{adaptive\_expansion} uses rollout confidence to decide which intermediate states to expand; and (e) newly generated nodes are scored and added to the graph, with $g$-costs updated and nodes reassigned to their depth-partitioned open lists $\mathcal{O}_d$ or the closed list $\mathcal{C}$ if their state has already been visited. Upon termination, the algorithm returns the lowest-cost plan found.

\section{Experiments}\label{sec:experiments}

We provide a large-scale evaluation of test-time inference methods for generative planning in classical AI planning domains. Our experiments address the following questions:
\begin{enumerate}[noitemsep, topsep=0pt, partopsep=0pt]
\item Does \textsc{OCLGen} produce shorter plans than baseline methods for a given compute budget while maintaining high completion rates?
\item Does \textsc{OCLGen} significantly increase the proportion of generated solutions that are optimal?
\item Which components of our method contribute most to its effectiveness?
\item To what extent is \textsc{OCLGen} suitable as the basis for a model self-improvement framework?
\end{enumerate}

\subsection{Test-time Inference Benchmark}\label{sec:experiments:test_time_inference}

\paragraph{Planning Domains.} We consider the following domains with varying challenges and levels of complexity:
\begin{itemize}[noitemsep, topsep=0pt]
\item \textbf{Blocksworld}: A classic domain where blocks must be rearranged into goal configurations. Optimal planning is NP-hard~\citep{SLANEY2001119}.
\item \textbf{Logistics}: A transportation domain where packages are delivered between locations using trucks and airplanes across multiple cities.
\item \textbf{Labyrinth}: A navigation domain on a grid where the agent can move between connected cells or shift entire rows/columns, with cells wrapping around edges.
\item \textbf{Sokoban}: A puzzle domain where an agent pushes boxes to target locations while navigating around walls. Solving Sokoban is known to be PSPACE-complete~\citep{culberson98sokoban}.
\end{itemize}

\paragraph{Training and Evaluation Setup.} For each domain, we train $\pi_\theta$ and $h_\phi$ on $10^5$ problem instances with suboptimal solutions generated by Fast Downward (LAMA-first configuration~\citep{richter2010lama}). Both models use the data augmentation described in Section~\ref{sec:base_model_improvements}. We evaluate on 1000 held-out test instances per domain. We use the GPT-2 \citep{radford2019language} architecture to implement $\pi_\theta$, and a smaller BERT encoder \cite{devlin2019bert} for $h_\phi$. Both models are trained using the standard cross-entropy loss. Final models are selected based on the checkpoint with the lowest validation loss. Further details on model architecture and training hyperparameters are provided in App.~\ref{app:training_detail}. 

\paragraph{Baselines.} We compare against the following baselines:
\begin{itemize}[noitemsep, topsep=0pt]
\item \textbf{MCTS (full rollouts)}: Monte Carlo Tree Search using $\pi_\theta$ for rollouts and with PUCT selection. We use $h_\phi$ to compute values of unsuccessful rollouts.
\item \textbf{MCTS (partial rollouts)}: MCTS with shorter rollouts, using $h_\phi$ to obtain a value for backpropagation from the last state of the rollout.
\item \textbf{OCL-Anytime A*}: Classical A* search using $h_\phi$ as the heuristic, run in anytime mode.
\item \textbf{OCL-GBFS}: Greedy best-first search using $h_\phi$ as the heuristic for node prioritization.
\item \textbf{Best-of-N}: Generates a set of plan sequences with $\pi_\theta$ given a runtime limit and returns the shortest valid one.
\item \textbf{FD-LAMA-anytime}: Fast Downward with LAMA configuration~\citep{richter2010lama} which returns the best solution within the time limit.
\item \textbf{FD-LAMA-first}: This baseline returns the first solution found by FD-LAMA-anytime. This method generated our data for generative model pretraining.
\item \textbf{FD-optimal}: Fast Downward with the LM-Cut heuristic (48h timeout), establishing reference sets of optimal plans for each domain.
\end{itemize}

All learned methods use the same generative and heuristic models, isolating the effect of the inference algorithm. A 10-minute timeout is given per problem (except for FD solvers). For \textsc{OCLGen}, we evaluate two different depth selection strategies \emph{uniform} and \emph{scan} (Section~\ref{sec:method:depth_selection}). We empirically determined an action confidence threshold $\tau_\text{conf}{=}0.95$ for all domains, except for Logistics where we use $\tau_\text{conf}{=}0.2$. More details on hyperparameter choices for inference are given in Appendix~\ref{app:evaluation_hyperparameters}.

\begin{table*}[htb]
\centering
\resizebox{\textwidth}{!}{%
\begin{tabular}{l|cc|cc|cc|cc}
\toprule
& \multicolumn{2}{c|}{\textbf{Blocksworld}} & \multicolumn{2}{c|}{\textbf{Logistics}} & \multicolumn{2}{c|}{\textbf{Labyrinth}} & \multicolumn{2}{c}{\textbf{Sokoban}} \\
\textbf{Method} \small{($t_\text{max}{=}$10min)} & Comp.[$\%$]  & Length & Comp. [$\%$] & Length & Comp. [$\%$] & Length & Comp. [$\%$] & Length \\
\midrule
\textbf{OCLGen} (uniform)               & \hfill 100.0 & \hfill 43.88 (± 0.68) & \hfill 100.0 & \hfill 155.83 (± 3.57) & \hfill 100.0 & \hfill  12.99 (± 0.13) & \hfill 99.9 & \hfill 128.67 (± 3.15)  \\
\textbf{OCLGen} (scan)                  & \hfill 100.0 & \hfill 44.10 (± 0.70) & \hfill 100.0 & \hfill 157.63 (± 3.51) & \hfill 100.0 & \hfill 13.00 (± 0.13) & \hfill 99.7 & \hfill 128.10 (± 3.15) \\
\hline
MCTS (full rollouts)                    & \hfill 100.0 & \hfill 54.56 (± 0.93) & \hfill 100.0 & \hfill 158.75 (± 3.52) & \hfill 100.0 & \hfill 15.81 (± 0.23) & \hfill 99.8 & \hfill 131.30 (± 3.23) \\
MCTS (partial rollouts)                 & \hfill 100.0 & \hfill 53.81 (± 0.91) & \hfill 29.3 & \hfill 40.31 (± 1.30) & \hfill 100.0 & \hfill 14.77 (± 0.19)  & \hfill 37.3 & \hfill  38.34 (± 0.93)  \\      
OCL-Anyt. A*                            & \hfill 71.1 & \hfill 37.07 (± 0.71) & \hfill 1.6 & \hfill 2.81 (± 0.70) & \hfill 100.0 & \hfill 13.04 (± 0.13)  & \hfill 33.5 & \hfill  37.30 (± 1.10) \\
OCL-GBFS                                & \hfill 89.5 & \hfill 65.23 (± 1.50)  & \hfill 1.3 & \hfill 2.46 (± 0.85) & \hfill 99.2  & \hfill 18.03 (± 0.43)  & \hfill 35.9 &  \hfill 48.38 (± 1.71)  \\
Best-of-N                               & \hfill 99.9 & \hfill 61.56 (± 1.03) & \hfill 100.0 &  \hfill 157.00 (± 3.47)  & \hfill 100.0 & \hfill 17.42 (± 0.25) & \hfill 98.8 & \hfill 132.25 (± 3.22) \\
\hline
FD-LAMA-anytime \small{($t_\text{max}{=}$10min)} & \hfill 100.0 & \hfill 45.05 (± 0.73) & \hfill 100.0 & \hfill 161.00 (± 3.57) & \hfill 100.0 & \hfill 19.77 (± 0.41) & \hfill 99.4 & \hfill 141.43 (± 3.84) \\
FD-LAMA-anytime \small{($t_\text{max}{=}$20min)} & \hfill 100.0 & \hfill 44.85 (± 0.72) & \hfill 100.0 & \hfill 160.93 (± 3.57) & \hfill 100.0 & \hfill 16.25 (± 0.32) & \hfill 100.0 & \hfill 143.43 (± 3.84) \\
FD-LAMA-first \small{($t_\text{max}{=}$20min)} & \hfill 100.0 & \hfill 79.43 (± 1.50) & \hfill 100.0 &  \hfill 165.32 (± 3.42) & \hfill 100.0 & \hfill 25.78 (± 0.50) & \hfill 100.0 & \hfill 149.05 (± 3.62) \\

\hline
FD-optimal \small{($t_\text{max}{=}$48h)} & \hfill 63.0 & \hfill 30.12 (± 0.48) & \hfill 16.9 & \hfill 27.31 (± 1.48) & \hfill 100.0 & \hfill 12.96 (± 0.12) & \hfill 48.9 & \hfill 47.03 (± 1.08)  \\
\bottomrule
\end{tabular}%
}
\caption{Benchmark on unseen problems (1000 per domain). Comp.: Completion (\%); Length: Mean plan length (± std. error).}
\label{tab:main_results}
\end{table*}

\paragraph{Plan Lengths.} Table~\ref{tab:main_results} presents completion rates and plan lengths across all domains. \textsc{OCLGen} achieves the best combination of completion rate and plan quality among all methods operating within the 10-minute budget per problem instance. Both \textsc{OCLGen} variants attain near-perfect completion rates (99.7--100\%) while producing consistently shorter plans than competing approaches. Notably, \textsc{OCLGen} reduces average plan length by 19.6\% compared to MCTS with full rollouts on Blocksworld and by 17.8\% on Labyrinth. While MCTS with partial rollouts and search-based methods (OCL-Anyt.\ A*, OCL-GBFS) achieve competitive results on simpler domains, they fail to scale to the more challenging Logistics and Sokoban benchmarks, with completion rates dropping to 1.3--37.3\%. Best-of-N sampling achieves high completion rates but consistently yields longer plans than \textsc{OCLGen} across all domains. The FD-optimal solver, even with a 48-hour time limit, solves only 16.9--63\% of instances on Blocksworld, Logistics, and Sokoban, underscoring the inherent difficulty of these domains. Finally, the \emph{uniform} and \emph{scan} selection strategies yield comparable performance, while \emph{uniform} achieves slightly shorter plans overall except for Sokoban.

\begin{table*}[ht]
\centering
\resizebox{\textwidth}{!}{%
\begin{tabular}{l|cc|cc|cc|cc}
\toprule
& \multicolumn{2}{c|}{\textbf{Blocksworld}} & \multicolumn{2}{c|}{\textbf{Logistics}} & \multicolumn{2}{c|}{\textbf{Labyrinth}} & \multicolumn{2}{c}{\textbf{Sokoban}} \\
\textbf{Method} \small{($t_\text{max}{=}$10min)} & Optimal  & Length (solved) & Optimal & Length (solved) & Optimal & Length (solved) & Optimal & Length (solved)\\
\midrule
\textbf{OCLGen} (uniform)   & \hfill  528~/~630 & \hfill 30.64 (630~/~630)  & \hfill 104~/~169 & \hfill 28.73 (169~/~169) & \hfill 988~/~1000  & \hfill 12.99 (1000~/~1000) & \hfill 377~/~489 & \hfill 48.45 (489~/~489)\\
\textbf{OCLGen} (scan)    & \hfill 522~/~630 & \hfill 30.67 (630~/~630)  & \hfill 99~/~169 & \hfill 28.99 (169~/~169) & \hfill 987~/~1000 & \hfill 13.00 (1000~/~1000) & \hfill 384~/~489  & \hfill 48.29 (489~/~489) \\
\hline
MCTS (full rollouts)            & \hfill 180~/~630 & \hfill 36.48 (630~/~630)  & \hfill  64~/~169 & \hfill 29.70 (169~/~169)  & \hfill 605~/~1000 & \hfill 15.81 (1000~/~1000)  & \hfill 291~/~489  & \hfill 49.52 (488~/~489) \\
MCTS (partial rollouts)         & \hfill 186~/~630 & \hfill 36.48 (630~/~630) & \hfill 64~/~169 & \hfill 26.23 (160~/~169) & \hfill 671~/~1000 & \hfill 14.77 (1000~/~1000) & \hfill 271~/~489 & \hfill 38.34 (373~/~489) \\      
OCL-Anyt. A*        & \hfill 336~/~630 & \hfill 31.25 (594~/~630) & \hfill 16~/~169 & \hfill 2.81 (16~/~169) & \hfill 975~/~1000 & \hfill 13.04 (1000~/~1000) & \hfill 293~/~489 & \hfill 36.55 (333~/~489) \\
OCL-GBFS            & \hfill 146~/~630 & \hfill 42.69 (626~/~630) & \hfill 12~/~169 & \hfill 3.46 (13~/~169) & \hfill 589~/~1000 & \hfill 18.03 (992~/~1000) & \hfill 160~/~489 & \hfill 43.85 (341~/~489) \\
Best-of-N    & \hfill 85~/~630 & \hfill 42.40 (630~/~630)  & \hfill 59~/~169 & \hfill 29.93 (169~/~169) & \hfill 444~/~1000 & \hfill 17.42 (1000~/~1000)  & \hfill 251~/~489 & \hfill 50.87 (484~/~489) \\
\hline

FD-LAMA-anytime \small{($t_\text{max}{=}$10min)} & \hfill  591~/~630 & \hfill  30.53 (630~/~630)  & \hfill  105~/~169 & \hfill  28.56 (169~/~169) & \hfill  459~/~1000 & \hfill  19.77 (1000~/~1000) & \hfill  421~/~489 & \hfill  47.84 (489~/~489) \\

FD-LAMA-anytime \small{($t_\text{max}{=}$20min)} & \hfill  601~/~630 & \hfill  30.48 (630~/~630)  & \hfill  106~/~169 & \hfill  28.51 (169~/~169) & \hfill  653~/~1000 & \hfill  16.25 (1000~/~1000) & \hfill  441~/~489 & \hfill  47.58 (489~/~489) \\

FD-LAMA-first  \small{($t_\text{max}{=}$20min)} & \hfill   62~/~630 & \hfill 52.45 (630~/~630) & \hfill  47~/~169 & \hfill 31.21 (169~/~169) & \hfill  249~/~1000 & \hfill 25.78 (1000~/~1000) & \hfill  121~/~489 & \hfill 58.08 (489~/~489) \\

\hline
FD-optimal  \small{($t_\text{max}{=}$48h)}& \hfill  630~/~630 & \hfill 30.12 (630~/~630)  & \hfill  169~/~169  & \hfill 27.31 (169~/~169) & \hfill  1000~/~1000 & \hfill 12.96 (1000~/~1000) & \hfill  489~/~489 & \hfill 47.03 (489~/~489) \\

\bottomrule
\end{tabular}%
}
\caption{Benchmark on subset of known optimal problems. Optimal: num. solved optimally; Length: Plan length (num. solved).}
\label{tab:results_optimal}
\end{table*}

\paragraph{Solution Optimality.} Table~\ref{tab:results_optimal} examines plan statistics on the subset of instances for which optimal solutions are known (solved by FD-optimal within 48h). \textsc{OCLGen} yields the most optimal solutions among all learning methods, solving 83.8\% (528/630) of Blocksworld, 61.5\% (104/169) of Logistics, 98.8\% (988/1000) of Labyrinth, and 77.1\% (377/489) of Sokoban with the \emph{uniform} variant. The \emph{scan} variant performs comparably, marginally improving the Sokoban rate to 78.5\% (384/489). This represents a 2.9$\times$ improvement over MCTS with full rollouts on Blocksworld and a 1.6$\times$ improvement on Labyrinth. Notably, \textsc{OCLGen}'s (\emph{uniform}) average plan lengths on solved instances closely approach those of the FD-optimal solver: within 1.7\% on Blocksworld (30.64 vs.\ 30.12), 5.2\% on Logistics (28.73 vs.\ 27.31), 0.2\% on Labyrinth (12.99 vs.\ 12.96), and 3.0\% on Sokoban (48.45 vs.\ 47.03). While OCL-Anyt.\ A* achieves competitive optimality rates on Blocksworld (336/630) and Labyrinth (975/1000), it fails to solve the majority of Logistics and Sokoban instances within the time limit. On this subset, FD-LAMA-anytime (10min) yields more optimal solutions in Blocksworld, Logistics, and Sokoban. However, these instances favor lower-complexity problems solvable by FD-optimal within 48 hours. Crucially, on the full test set---which spans higher-complexity problems---\textsc{OCLGen} scales significantly better, producing shorter plans than FD-LAMA-anytime across all four domains (Table~\ref{tab:main_results}). Moreover, we later show that our method can distill its search results back into the model, surpassing FD-LAMA-anytime even on this subset after recursive self-improvement (Section~\ref{sec:self_improvement}). Overall, these results demonstrate that \textsc{OCLGen} significantly increases the percentage of optimal solutions while maintaining high completion rates.

\begin{figure*}[t]
    \centering
    \begin{subfigure}[b]{0.24\textwidth}
        \centering
        \includegraphics[width=\textwidth]{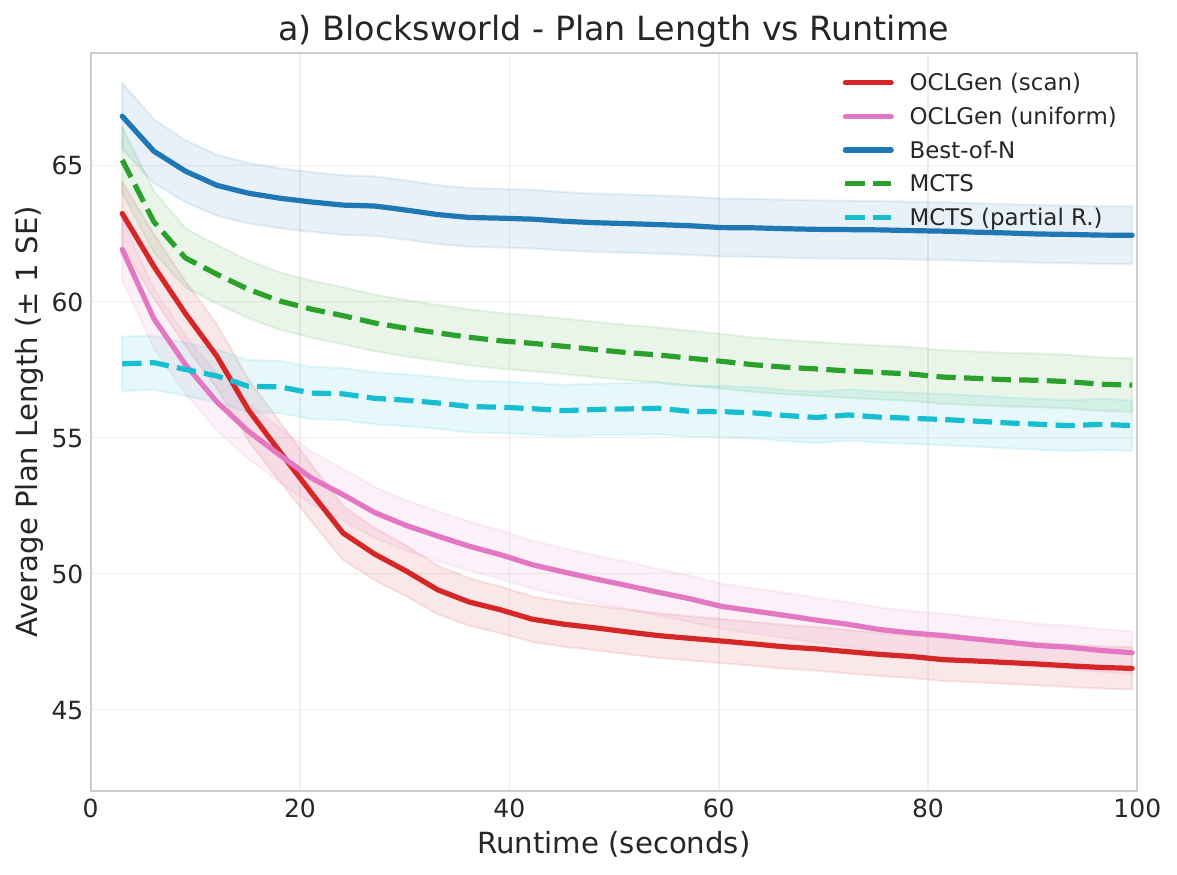}
        \caption{Blocksworld}
        \label{fig:time_blocksworld}
    \end{subfigure}
    \hfill
    \begin{subfigure}[b]{0.24\textwidth}
        \centering
        \includegraphics[width=\textwidth]{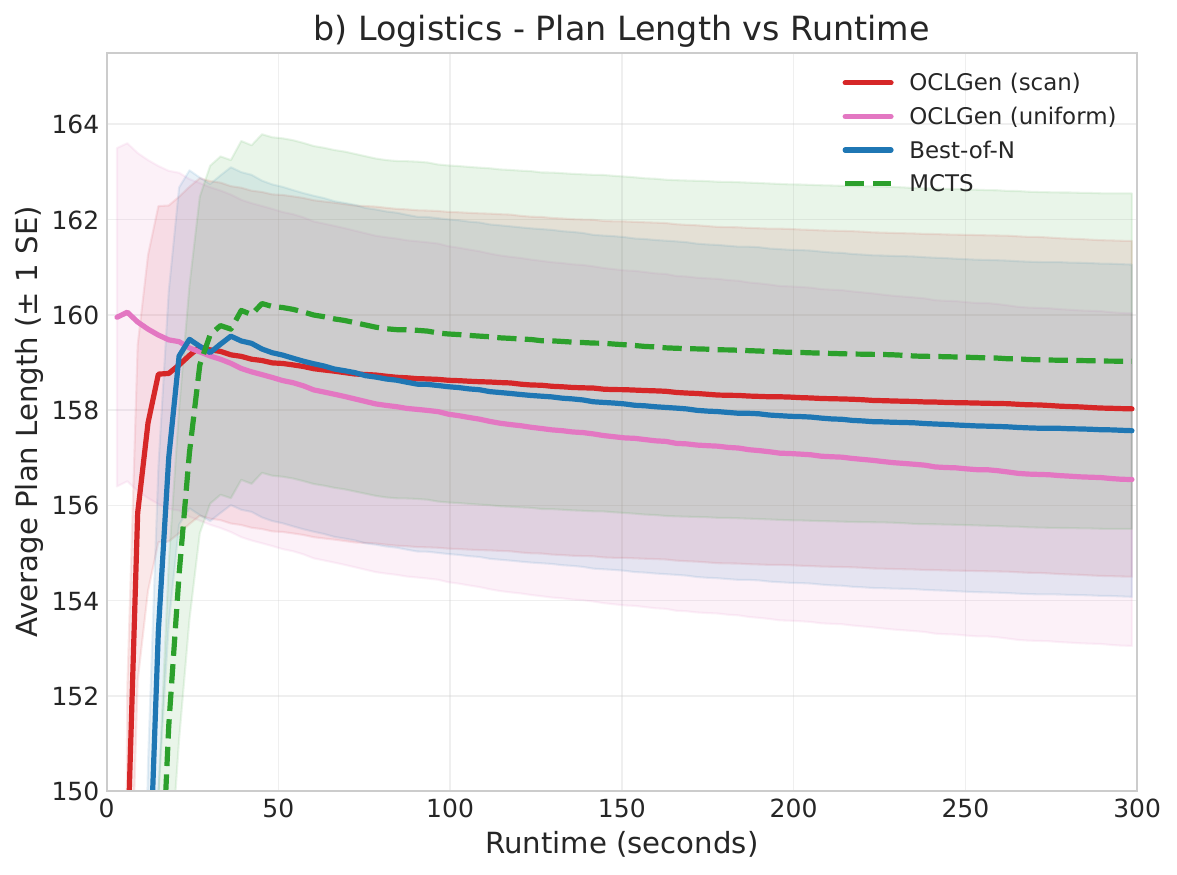}
        \caption{Logistics}
        \label{fig:time_logistics}
    \end{subfigure}
    \hfill
    \begin{subfigure}[b]{0.24\textwidth}
        \centering
        \includegraphics[width=\textwidth]{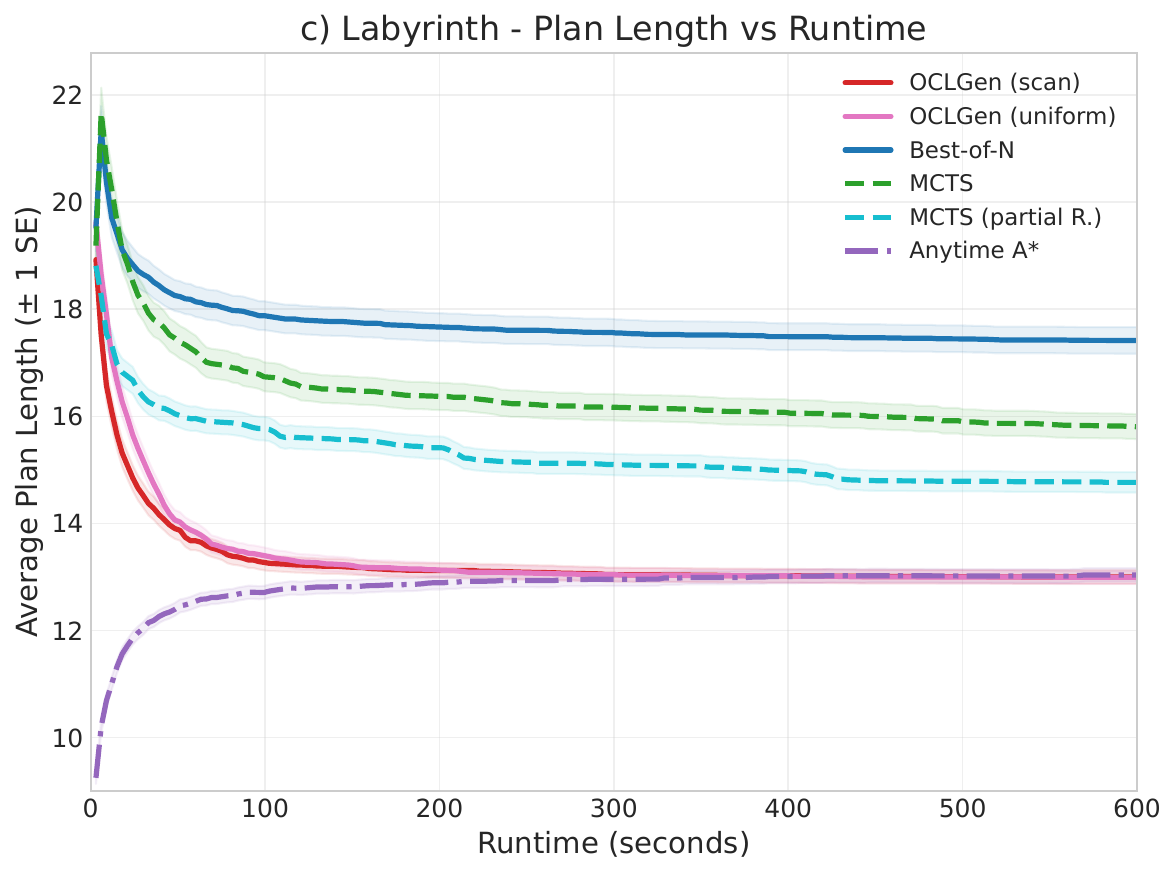}
        \caption{Labyrinth}
        \label{fig:time_labyrinth}
    \end{subfigure}
    \hfill
    \begin{subfigure}[b]{0.24\textwidth}
        \centering
        \includegraphics[width=\textwidth]{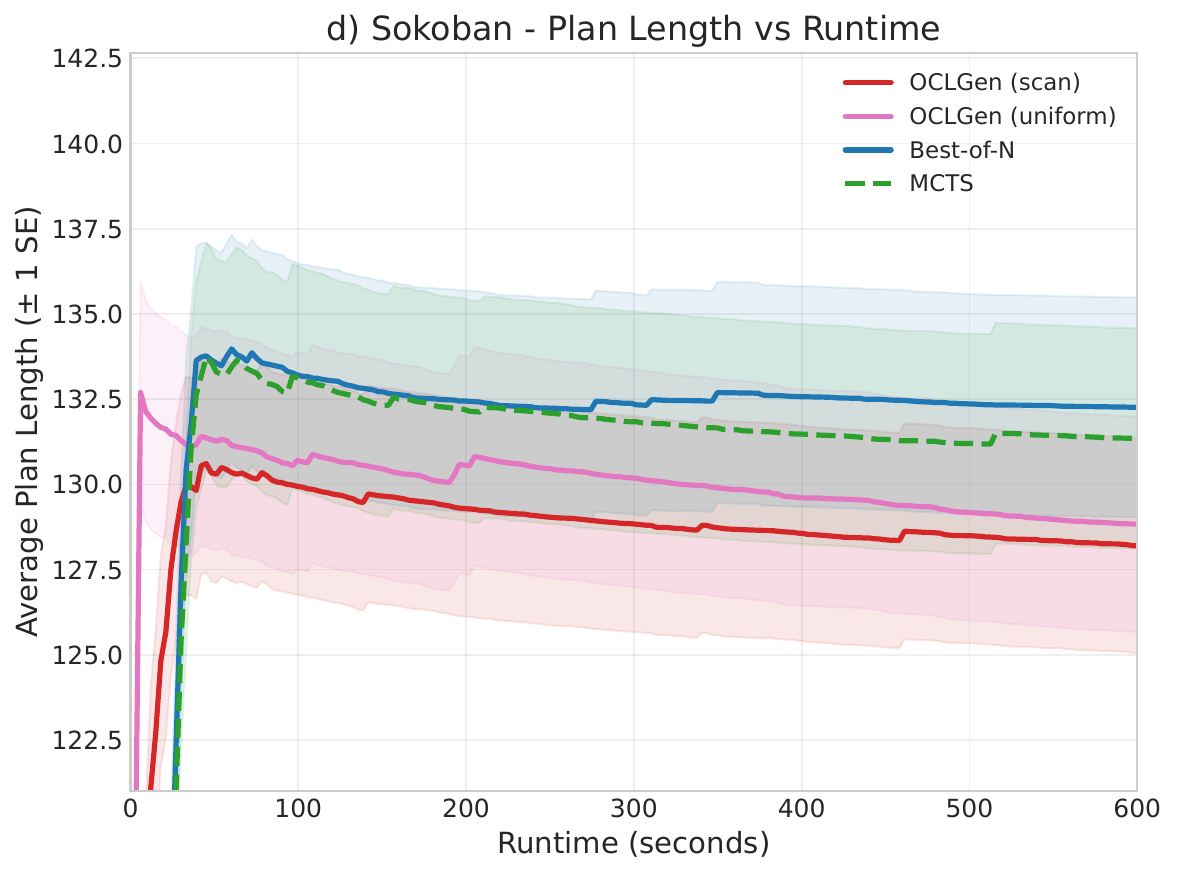}
        \caption{Sokoban}
        \label{fig:time_sokoban}
    \end{subfigure}
    \caption{Plan length over time across all domains. \textsc{OCLGen} rapidly converges to shorter plans compared to baseline methods.}
    \label{fig:length_over_time}
\end{figure*}

\begin{figure*}[ht]
    \centering
    \begin{subfigure}[b]{0.24\textwidth}
        \centering
        \includegraphics[width=\textwidth]{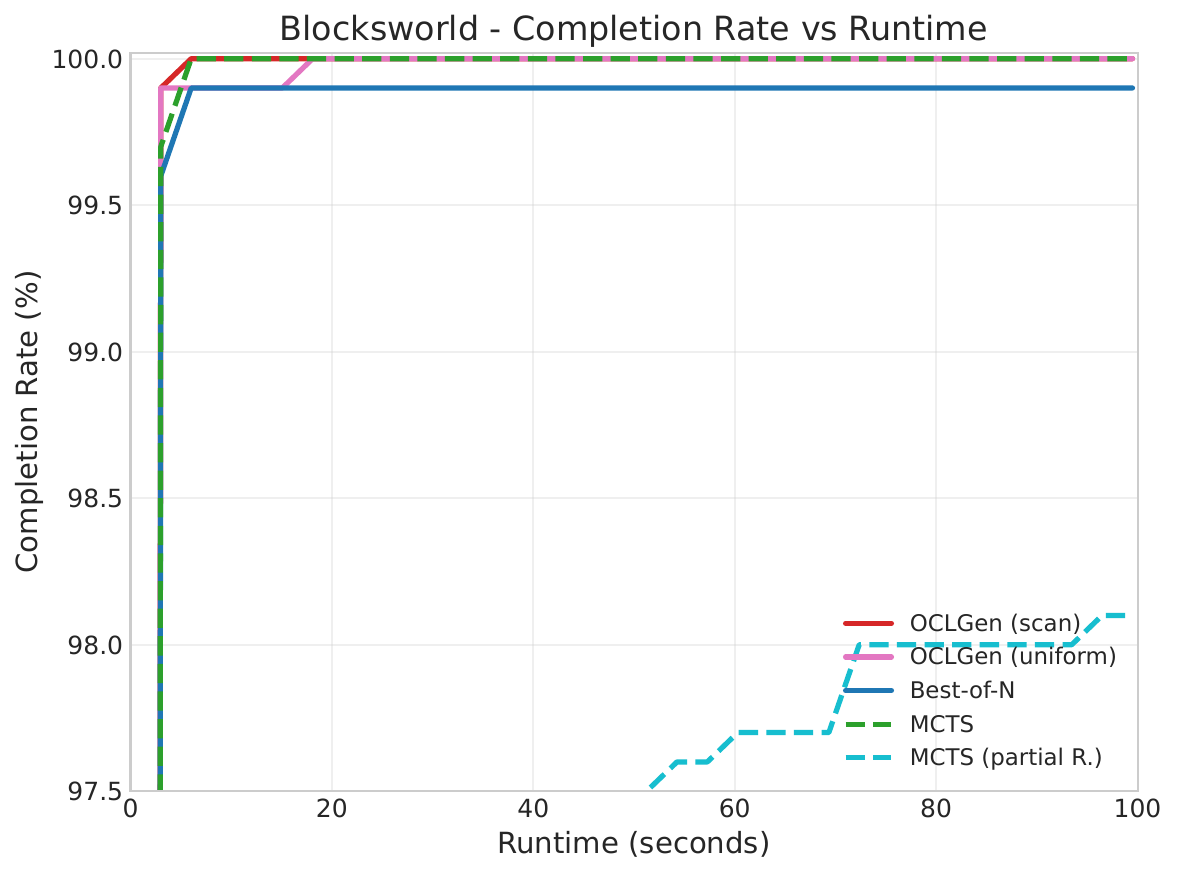}
        \caption{Blocksworld}
        \label{fig:time_blocksworld_completion}
    \end{subfigure}
    \hfill
    \begin{subfigure}[b]{0.24\textwidth}
        \centering
        \includegraphics[width=\textwidth]{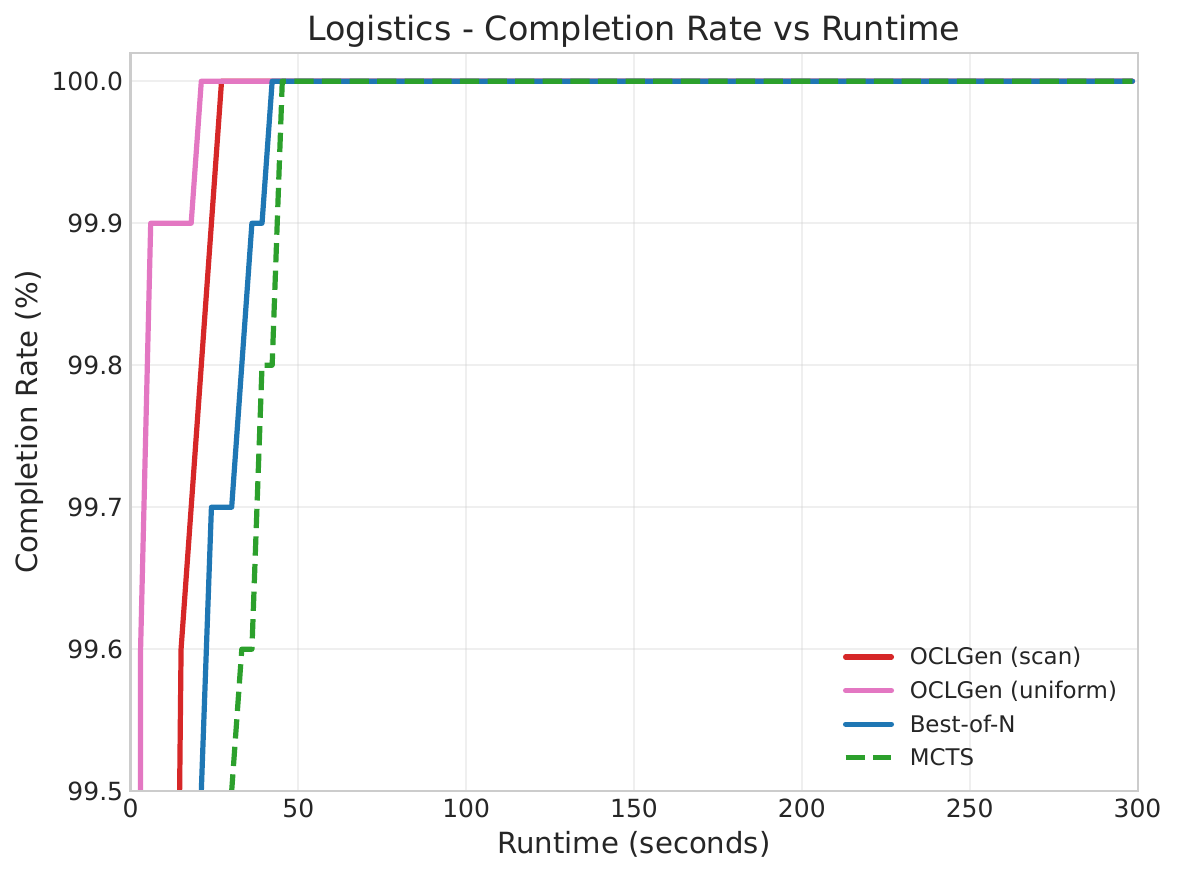}
        \caption{Logistics}
        \label{fig:time_logistics_completion}
    \end{subfigure}
    \hfill
    \begin{subfigure}[b]{0.24\textwidth}
        \centering
        \includegraphics[width=\textwidth]{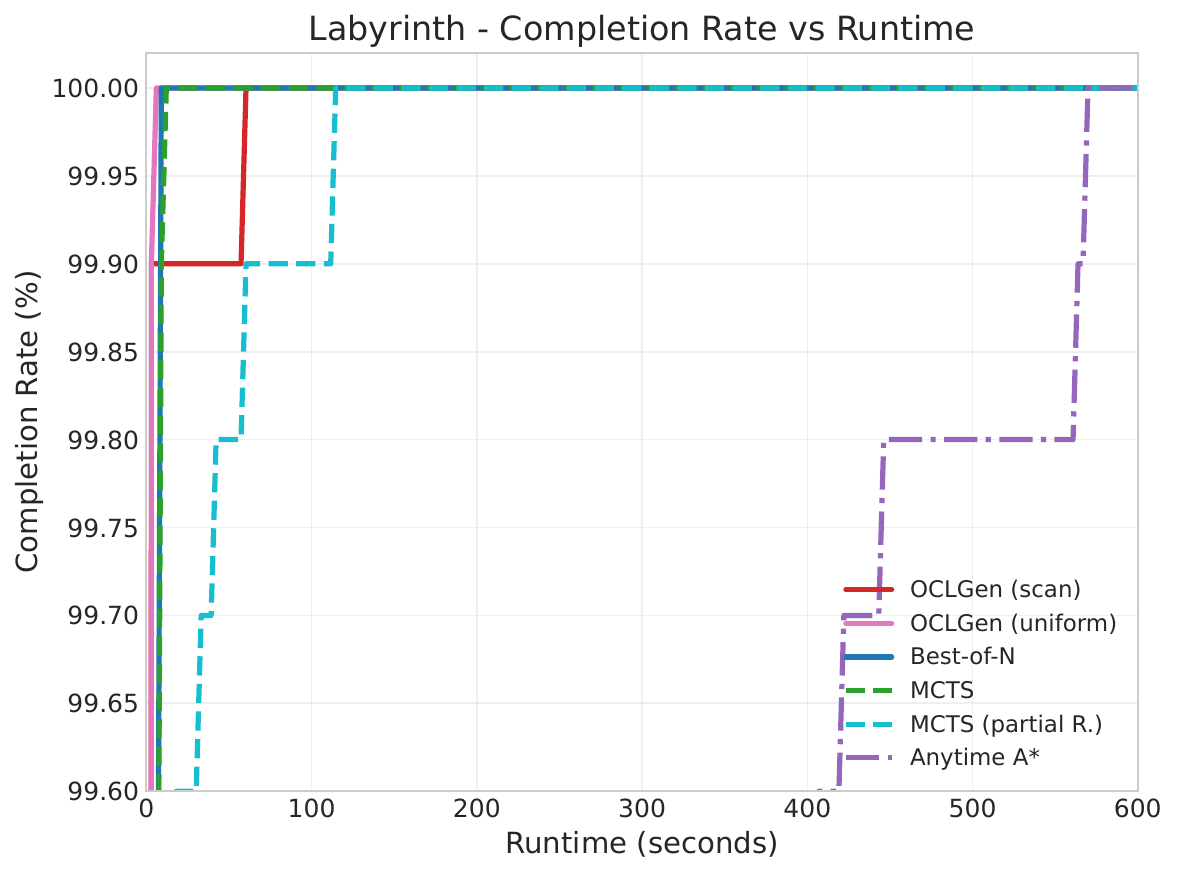}
        \caption{Labyrinth}
        \label{fig:time_labyrinth_completion}
    \end{subfigure}
    \hfill
    \begin{subfigure}[b]{0.24\textwidth}
        \centering
        \includegraphics[width=\textwidth]{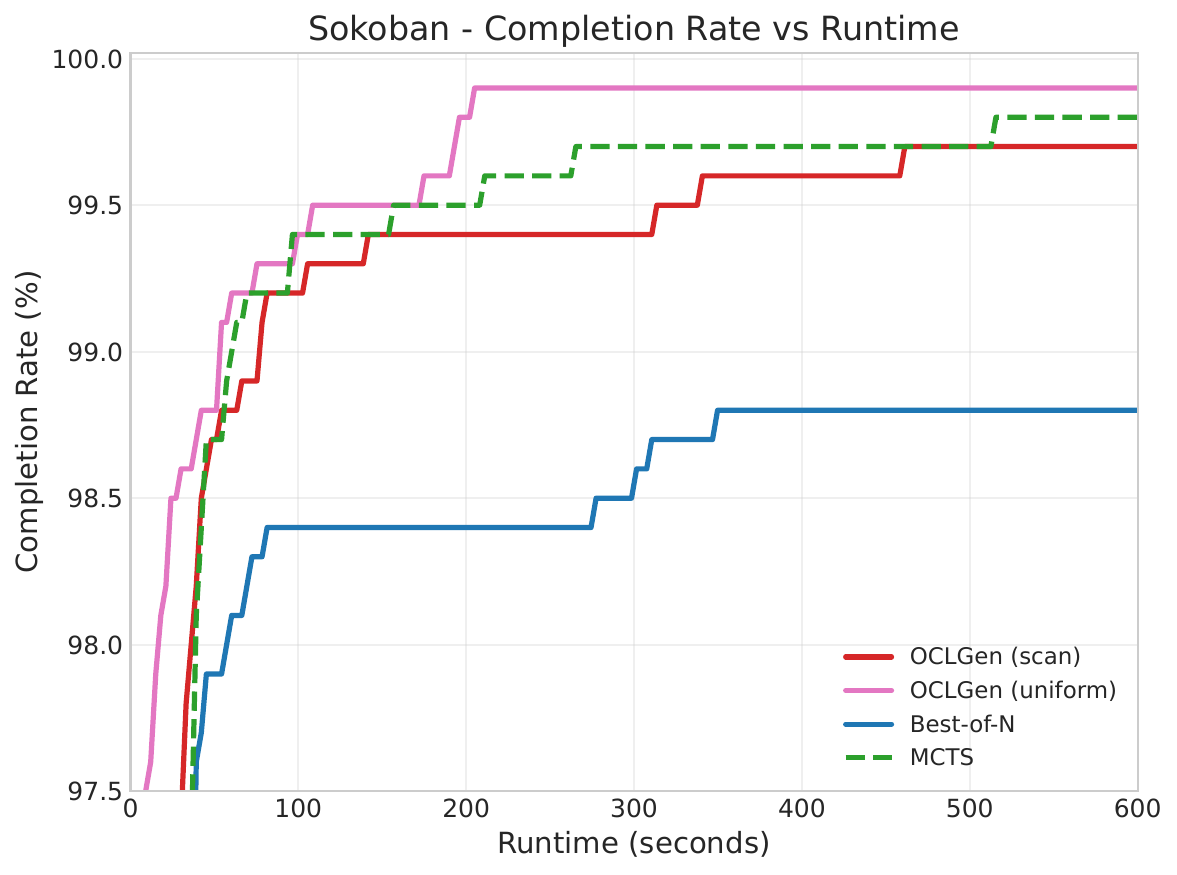}
        \caption{Sokoban}
        \label{fig:time_sokoban_completion}
    \end{subfigure}
    \caption{Completion rate over time across all domains.}
    \label{fig:completion_rate_over_time}
\end{figure*}

\paragraph{Plan Quality Convergence.} Figure~\ref{fig:length_over_time} shows the average plan length and completion rates against runtime. Note that average plan lengths initially increase as shorter problems tend to be solved first. Across all domains, \textsc{OCLGen} achieves shorter plans significantly faster than baseline methods. On Blocksworld, it reaches an average plan length of 50 steps within 30 seconds, a quality that MCTS and Best-of-N fail to match even after 5 minutes of runtime. Similarly, in Labyrinth, \textsc{OCLGen} converges to near-optimal lengths ($\sim$13 steps) in under 200 seconds, while MCTS requires the full 600-second budget to plateau at a substantially higher length. This pattern holds in Logistics and Sokoban, where \textsc{OCLGen} consistently achieves lower plan lengths after completion rates stabilized. The steeper descent of \textsc{OCLGen}'s curves indicates more effective use of compute: each additional second of runtime yields greater plan quality improvements compared to the baselines. Importantly, as shown in Figure~\ref{fig:completion_rate_over_time}, this improved plan quality does not come at the cost of completion rate. \textsc{OCLGen} reaches near-perfect completion rates within seconds on Blocksworld, Labyrinth, and Logistics, and remains competitive with MCTS on Sokoban.

\begin{table}[t]
\centering
\resizebox{\columnwidth}{!}{%
\begin{tabular}{l|cc|cc}
\toprule
& \multicolumn{2}{c|}{\textbf{Logistics}} & \multicolumn{2}{c}{\textbf{Sokoban}} \\
\textbf{Method} \small{($t_\text{max}{=}$10min)} & Comp.[$\%$] & Length & Comp.[$\%$] & Length \\
\midrule
\textbf{OCLGen} (uniform) & \hfill 100.0 & \hfill 374.85 (± 4.71) & \hfill 93.6 & \hfill 348.94 (± 3.66) \\
Best-of-N [$N_\text{max}{=}\infty$] & \hfill 98.5 & \hfill 376.09 (± 3.59) & \hfill 90.8 & \hfill 364.27 (± 4.19)  \\
\bottomrule
\end{tabular}%
}
\caption{Completion rate and plan length statistics of \textsc{OCLGen} (uniform) and Best-of-N on unseen problems for which the training data generation with FD-LAMA-first ($t_\text{max}{=}20$min) failed (number of problems:  66 for Logistics and 250 for Sokoban). Comp.: Completion (\%); Length: Mean plan length (± std. error).}
\label{tab:results_subset}
\end{table}

\subsection{Generalization Beyond Training Data}\label{sec:generalization_lama_failed}
Table~\ref{tab:results_subset} evaluates \textsc{OCLGen} (uniform) on problem instances where the training data planner (FD-LAMA with a 20-minute budget) failed to find a solution. \textsc{OCLGen} solves 100\% of these challenging Logistics instances and 93.6\% of Sokoban instances. This indicates that the generative model learns transferable strategies rather than merely overfitting the training data. Furthermore, \textsc{OCLGen} consistently outperforms Best-of-N sampling on these difficult instances, improving completion rates by 1.5 percentage points on Logistics and 2.8 percentage points on Sokoban, while also producing shorter plans (374.85 vs.\ 376.09 on Logistics; 348.94 vs.\ 364.27 on Sokoban).

\subsection{Ablations}

To evaluate the contribution of each component in \textsc{OCLGen}, we systematically ablate three key design choices across all four domains: (1) \textit{depth partitioning}, (2) \textit{adaptive expansion}, and (3) the \textit{percentile-based cost-to-go estimator}. For each ablation, we remove or replace a single component while keeping all other settings fixed, using \textsc{OCLGen} (uniform) as the base configuration. Table~\ref{tab:ablations} reports completion rates and plan lengths across 1000 test instances per domain, again given a 10-min runtime limit. Each component provides incremental improvements that collectively drive the full method's performance.

\begin{table}[t]
\centering
\resizebox{\columnwidth}{!}{
\begin{tabular}{l|l|ccc}
\toprule
\textbf{Domain} & \textbf{Metric} & \textbf{w/o Depth} & \textbf{w/o Adaptive} &  \textbf{w/o Percentile} \\
 & & \textbf{selection}  & \textbf{expansion} & \textbf{estimate (mode)} \\
\midrule
\multirow{3}{*}{\textbf{Blocksworld}} & Optim. & 366~/~630 & 496~/~630 & 486~/~630 \\
& Comp.[\%] & 100.0 & 100.0 & 100.0 \\
 & Length & 48.81 (± 0.87)  & 44.35 (± 0.69) & 44.68 (± 0.70) \\
\midrule
\multirow{3}{*}{\textbf{Logistics}} & Optim. & 78~/~169 & 98~/~169 & 98~/~169  \\
& Comp.[\%] & 99.9 & 100.0 & 100.0 \\
 & Length & 159.23 (± 3.52) & 155.82 (± 3.47) & 156.11 (± 3.48) \\
\midrule
\multirow{3}{*}{\textbf{Labyrinth}} & Optim. & 979~/~1000 & 983~/~1000  & 980~/~1000  \\
& Comp.[\%] & 100.0 & 100.0 & 100.0 \\
 & Length & 13.08 (± 0.15) & 13.02 (± 0.13) & 13.03 (± 0.13) \\
\midrule
\multirow{3}{*}{\textbf{Sokoban}} & Optim. & 353~/~489 & 341~/~489 & 350~/~489  \\
& Comp.[\%] & 99.7 & 100.0 & 99.8 \\
 & Length & 129.39 (± 3.17) &  129.73 (± 3.15) & 129.52 (± 3.17) \\
\bottomrule
\end{tabular}
}
\caption{Ablation study across problem domains. Comp.: Completion rate (\%); Length: Plan length (mean $\pm$ std. error).}
\label{tab:ablations}
\end{table}

\paragraph{Impact of Depth Partitioning.}
Table~\ref{tab:ablations} shows that removing depth-based selection leads to notably longer plans, particularly on Blocksworld (48.81 vs.\ 43.88) and Logistics (159.23 vs.\ 155.83). Without depth partitioning, the search tends to over-exploit deep branches, reducing the diversity of explored trajectories and ultimately yielding suboptimal solutions.

\paragraph{Confidence-based Adaptive Expansion.}
The adaptive expansion mechanism dynamically decides when to expand the search tree based on rollout confidence. Removing this component yields comparable completion rates and overall longer plans (e.g. 44.35 vs.\ 43.88 on Blocksworld). The adaptive threshold allows \textsc{OCLGen} to allocate more computational effort to uncertain regions of the search space while quickly committing to high-confidence trajectories, leading to an improvement in plan quality.

\paragraph{Influence of Heuristic Point Estimator.}
We compare our percentile-based cost-to-go estimate against using the mode of the predicted distribution. Table~\ref{tab:ablations} shows that replacing the percentile estimate with the mode results in marginally longer plans across all domains. While the differences are modest, the percentile-based estimator yields more optimistic cost-to-go estimates that counteract overestimation bias, improving plan quality on Blocksworld (43.88 vs.\ 44.68) and Logistics (155.83 vs.\ 156.11).

\subsection{OCLGen for Model Self-Improvement}\label{sec:self_improvement}

We evaluate the use of \textsc{OCLGen} for recursive model self-improvement. Given a random subset of training problems, we run our search algorithm for a fixed runtime budget on each problem instance to compute improved plans, then finetune both $\pi_\theta$ and $h_\phi$ on this data. This procedure is applied recursively, using updated model weights from the previous iteration to generate improved plans on newly sampled problems. We compare \textsc{OCLGen} (uniform) and MCTS (full rollouts) for $n_\text{loop}{=}3$ iterations of self-improvement in Blocksworld and Sokoban. At each iteration, we sample 3000 problems and run the search method on each of these given a time limit of 3 minutes in Blocksworld and 5 minutes in Sokoban (further details in Appendix~\ref{app:self_improvement_hyperparameters}).
\begin{table}[t]
\centering
\resizebox{\columnwidth}{!}{
\begin{tabular}{c|c|cc|cc}
\toprule
& & \multicolumn{2}{c|}{\textbf{OCLGen} (uniform)} & \multicolumn{2}{c}{\textbf{MCTS} (full rollouts)} \\
& $i_\text{loop}$ & Comp.[\%] & Length & Comp.[\%] & Length \\
\midrule
\multirow{4}{*}{\rotatebox[origin=c]{90}{\textbf{Blocksw.}}} & 0 & \hfill 99.9 & \hfill 69.51 (± 1.24) & \hfill 99.9 & \hfill 69.51 (± 1.24) \\
& 1 & \hfill 99.2 & \hfill 47.70 (± 0.84) & \hfill 99.9 & \hfill 62.20 (± 1.18) \\
& 2 & \hfill 99.6 & \hfill 41.81 (± 0.58) & \hfill 99.9 & \hfill 57.30 (± 1.06) \\
& 3 & \hfill 99.3 & \hfill 41.40 (± 0.57) & \hfill 99.9 & \hfill 53.15 (± 0.98) \\
\hline\hline
\multirow{4}{*}{\rotatebox[origin=c]{90}{\textbf{Sokob.}}} & 0 & \hfill 95.8 & \hfill 135.00 (± 3.38) & \hfill 95.8 & \hfill 135.00 (± 3.38) \\
& 1 & \hfill 95.2 & \hfill 128.41 (± 3.25) & \hfill 95.1 & \hfill 128.89 (± 3.24) \\
& 2 & \hfill 93.5 & \hfill 123.93 (± 3.17) & \hfill 93.9 & \hfill 125.94 (± 3.18) \\
& 3 & \hfill 93.4 & \hfill 121.88 (± 3.11) & \hfill 93.9 & \hfill 124.48 (± 3.16) \\

\bottomrule
\end{tabular}
}
\caption{Self-improvement plan statistics on the test dataset (1000 samples) using best-of-N sampling ($N{=}10$). Comp.: Completion rate (\%); Length: Plan length (mean $\pm$ std. error).}
\label{tab:improvement_loop}
\end{table}

Table~\ref{tab:improvement_loop} reports the test set evaluation of the finetuned generative models after each self-improvement iteration using Best-of-N sampling ($N{=}10$). Both methods maintain near-full completion on Blocksworld, but \textsc{OCLGen} consistently produces shorter plans, providing a higher quality training signal. On Sokoban, a similar pattern is observed, although completion rates drop for both methods, presumably because finetuning on improved plans reduces policy entropy, decreasing diversity within the Best-of-N batch.

Table~\ref{tab:improvement_final} details the test set evaluation of our final models, which utilize test-time search after three rounds of self-improvement. On Blocksworld, \textsc{OCLGen} yields significantly shorter plans (40.74 vs.\ 44.89) and achieves 100\% optimal solutions (vs.\ 71.4\% for MCTS). On Sokoban, our method produces shorter plans (123.60 vs.\ 125.06) and achieves 94.7\% optimal solutions (vs.\ 81.8\% for MCTS), while recovering from the completion rate drop observed with the model alone to achieve 99.8\% completion. Notably, on the problem subsets with known optimal solutions, self-improvement with \textsc{OCLGen} surpasses the FD-LAMA-anytime solver (601/630 on Blocksworld and 441/489 on Sokoban at a 20-minute budget; Table~\ref{tab:results_optimal}). These results demonstrate the suitability of \textsc{OCLGen} for efficient model self-improvement.

\begin{table}[t]
\centering
\resizebox{\columnwidth}{!}{
\begin{tabular}{c|l|c|c}
\toprule
& & \textbf{OCLGen} (uniform) & \textbf{MCTS} (full rollouts) \\
\midrule
\multirow{4}{*}{\rotatebox[origin=c]{90}{\textbf{Blocksw.}}} & Plan Length & 40.74 (± 0.55) & 44.89 (± 0.70) \\
& Compl. [\%] & 100.0 & 100.0 \\ 
\cline{2-4}
& Plan length (optimal) & 30.12 (± 0.48) & 31.14 (± 0.53) \\
& Optimal & 630~/~630 (100.0\%) & 450~/~630 (71.4\%) \\
\hline\hline
\multirow{4}{*}{\rotatebox[origin=c]{90}{\textbf{Sokob.}}} & Plan Length & 123.60 (± 3.02) &  125.06 (± 3.08) \\
& Compl. [\%] & 99.8 & 99.3   \\ 
\cline{2-4}
& Plan length (optimal) & 47.17 (± 1.09) & 47.55 (± 1.10) \\
& Optimal & 463~/~489 (94.7\%) & 400~/~489 (81.8\%) \\
\bottomrule
\end{tabular}
}
\caption{Results on Blocksworld and Sokoban with \textsc{OCLGen} and MCTS ($t_\text{max}{=}$10min) after $n_\text{loop}=3$ iteration of model self-improvement on the test sets from Table~\ref{tab:main_results} (1000 samples). Comp.: Completion rate (\%); Length: Plan length (mean $\pm$ std. error). Optimal: number solved optimally.}
\label{tab:improvement_final}
\end{table}

\subsection{Accuracy of Heuristic Model}
Table~\ref{tab:mae_comparison} reports mean absolute error (MAE) statistics for Blocksworld and Sokoban before and after self-improvement. The MAE is computed on the subset of instances with known optimal solutions, using either the \emph{mode} or \emph{k-th percentile} to derive scalar values from the predicted cost-to-go distributions ($k=3$ and $k=10$ for Blocksworld and Sokoban, respectively). As expected, percentile-based estimates yield lower MAE than mode-based estimates for all base models ($n_\text{loop}=0$), reducing the gap to optimality. This behavior holds consistently across all four domains (see Table~\ref{tab:mae_me_comparison} in the Appendix).

Self-improvement via \textsc{OCLGen} ($n_\text{loop}=3$) substantially reduces the model's MAE -- for example, from 4.07 to 0.66 on Blocksworld and from 3.31 to 2.51 on Sokoban (percentile method). This confirms that \textsc{OCLGen} not only discovers better solutions across self-improvement rounds, but also successfully refines the heuristic model.

\begin{table}[t]
\centering
\resizebox{\columnwidth}{!}{
\begin{tabular}{lcccc}
\toprule
 & \multicolumn{2}{c}{\textbf{Blocksworld}} &  \multicolumn{2}{c}{\textbf{Sokoban}} \\
\midrule
$n_\text{loop}$ & 0 & 3 & 0 & 3 \\
\midrule
Mode & 8.68 (± 0.24) & 0.44 (± 0.02) & 5.28 (± 0.17) & 3.05 (± 0.11) \\
Perc. & 4.07 (± 0.12) & 0.66 (± 0.02) & 3.31 (± 0.12) & 2.51 (± 0.08) \\
\bottomrule
\end{tabular}}
\caption{MAE (mean absolute error) $\pm$ std. error of heuristic models on the subset of known optimal solutions. Values are reported for both mode-based and percentile-based heuristics.}\label{tab:mae_comparison}
\end{table}

\section{Conclusion}\label{sec:conclusion}

We present \textsc{OCLGen}, a test-time search algorithm adapting classical OCL search to autoregressive generative planning models. Key innovations include depth-partitioned selection, partial rollouts with adaptive expansion, and learned distributional heuristics. Across four domains, \textsc{OCLGen} delivers near-optimal plans at high completion rates, outperforming neurosymbolic and classical baselines on the full test set. Finally, on the known-optimal subsets, recursive self-improvement yields a 100\% optimality rate on Blocksworld and 94.7\% on Sokoban.

\paragraph{Future Work.} Our depth selection strategies are currently uninformed, following fixed schedules. Adaptive strategies that concentrate search effort on the most promising depths may yield further gains. Characterizing when recursive self-improvement provably converges to optimal solutions remains an open theoretical question. Scaling to larger problems and generalizing to object counts or grid sizes outside the training distribution remain open challenges, as does transfer to new domains. Finally, our method assumes suboptimal initial solutions for pretraining. Extending \textsc{OCLGen} to support efficient policy improvement without prior data represents a compelling future direction.

\clearpage
\section*{Impact Statement}
This paper presents work whose goal is to advance the field of Machine
Learning. There are many potential societal consequences of our work, none
which we feel must be specifically highlighted here.

\bibliography{main}
\bibliographystyle{icml2026}

\newpage
\newpage
\appendix
\onecolumn

\section{Datasets}
\label{app:datasets}

We construct dedicated datasets for each of the four planning domains evaluated in this work. Each dataset comprises 100k training instances, 1k validation instances for model selection, and 1k held-out test instances. All problem instances are unique and paired with reference solutions obtained using Fast Downward~\cite{HELMERT2003219} in the LAMA-first configuration with a 20-minute timeout. Instances that could not be solved within this time limit were excluded from the training datasets.

\paragraph{Blocksworld}

We generate Blocksworld instances containing 3 to 25 blocks using the generator from~\cite{seipp2022pddlgen}. All generated instances were solvable by FD-LAMA within the 20-minute budget. The distribution of instances follows a logarithmic scaling with respect to the number of blocks, as the limited number of unique configurations for small block counts precludes a uniform distribution.

\paragraph{Logistics}

Our Logistics dataset covers a broad range of problem sizes: 1--50 cities with 1--5 locations each, 1--50 packages, 1--10 airplanes, and one truck per city. Instances are sampled uniformly over all valid combinations of these parameters. Fast Downward solved 99.97\% of generated instances within the time limit; additional instances were generated to ensure the target dataset sizes. 

\paragraph{Labyrinth}

Labyrinth was introduced in IPC 2023~\cite{taitler2023ipc}. We generate a plan dataset with $3 \times 3$ and $4 \times 4$ grids, which remain challenging while ensuring sufficient training coverage using FD-LAMA. Problem instances were generated using the official IPC 2023 generator~\cite{eifler2023labyrinth}.

\paragraph{Sokoban}

The Sokoban PDDL domain appeared in IPC 2008 and 2011. We constrain our instances to grid sizes between $5 \times 5$ and $14 \times 14$, with 1--10 boxes and up to 10 walls. We use the generator from~\cite{seipp2022pddlgen}.

\clearpage
\section{Baselines}

\paragraph{Monte Carlo Tree Search.}
We implemented Monte Carlo Tree Search using a PUCT-style selection policy similar to \citet{silver2016mastering}. Specifically, we maintain separate Q-values: $Q_\text{opt}$ for expected plan length (normalized by plan lengths of neighboring actions) following a transition, and $Q_\text{sat}$ for the expected probability of finding a solution. The overall Q-value is computed as a convex combination of $Q_\text{opt}$ and $Q_\text{sat}$ using a mixing coefficient $\alpha$. To compute the policy prior $P(s,a)$, we determine action probabilities from token sequences by computing the geometric mean of token probabilities for individual action sequences, then normalize over the set of possible actions at a state. The selection policy is $a^* = \arg \max_{a\in\mathcal{A}} \left[ \alpha\, Q_\text{sat}(s,a) + (1-\alpha)\, Q_\text{opt}(s,a) + c_\text{PUCT}\, P(s,a)\, \frac{\sqrt{\sum_b N(s,b)}}{1 + N(s,a)} \right]$. For all experiments, we set $\alpha=0.1$ and $c_\text{PUCT}=1.0$. We also incorporate progressive widening \citep{coulom07}, which significantly improved performance. For MCTS with partial rollouts, we use a maximum token limit of 50. We use the heuristic model $h_\phi$ to estimate the cost-to-go of unsuccessful rollouts.

\paragraph{Anytime A*.}
We implement a version of Anytime A* using our learned heuristic $h_\phi$ to guide the search. Similar to our method, we use the lower percentile to obtain heuristic point estimates from distributions. At every step, we select the node with the lowest $f$ value from the open list, where $f=g+h$. The search does not stop once a goal is found but continues until the runtime limit is exhausted or no unvisited nodes remain in the open list.

\paragraph{Greedy Best-First Search.}
This method is similar to Anytime A* but purely uses the learned heuristic $h_\phi$ to guide the search (i.e., $f=h$). Again, the search continues after finding a goal until the runtime limit is exhausted or no unvisited nodes remain.

\paragraph{Best-of-N.}
This method involves repeated sampling from the generative model $\pi_\theta$. For all experiments, we use a softmax temperature of $T_\text{softmax}=1$ and sample a batch of 10 plans. The generated plans are then validated for syntactic and semantic correctness, and we report the shortest valid plan in the batch. For the experiments, we terminate once the time limit of $t_\text{max}{=}10$ minutes is reached. Table~\ref{tab:number_of_rollout_time_limit_best_of_N} presents the number of candidate plans per domain generated by Best-of-N within this runtime budget.

\begin{table}[h]
\centering
\begin{tabular}{lc}
\toprule
\textbf{Domain} & \textbf{Number of Generated Plans} ($t_\text{max}{=}10$min) \\
\midrule
Blocksworld &  24768.9 ± 11816.0 \\
Logistics &  7243.4 ± 14927.1 \\
Labyrinth &  10953.1 ± 16137.1 \\ 
Sokoban & 2186.5 ± 4266.4  \\
\bottomrule
\end{tabular}
\caption{Number of generated plans (mean ± std. dev.) with Best-of-N for the experiments in Table~\ref{tab:main_results}}
\label{tab:number_of_rollout_time_limit_best_of_N}
\end{table}

\clearpage
\section{Training Details}\label{app:training_detail}

We train domain-specific policy and heuristic cost models on 4 compute instances, each equipped with 8 NVIDIA A100 GPUs (40GB VRAM). Table~\ref{tab:pre_hyperparameters} shows the transformer architecture and training hyperparameters for the generative model. Table~\ref{tab:pre_hyperparameters_heuristic_model} shows corresponding hyperparameters for the heuristic cost model. We use the same model architectures across all domains. 

\begin{table}[h]
\centering
\begin{tabular}{lc}
\toprule
\textbf{Hyperparameter} & \textbf{Value} \\
\midrule
\multicolumn{2}{l}{\textit{Architecture}} \\
\quad Type & Transformer Decoder (GPT2)  \\
\quad Transformer blocks ($n_\text{layer}$) & 12 \\
\quad Attention heads ($n_\text{head}$) & 12 \\
\quad Embedding dimension ($n_\text{embd}$) & 768 \\
\quad Feedforward dimension ($n_\text{inner}$) & 3072 \\
\midrule
\multicolumn{2}{l}{\textit{Optimization}} \\
\quad Learning rate & 5e-5 \\
\quad Scheduler & Cosine \\
\quad Warmup steps & 1000 \\
\quad Optimizer & AdamW \\
\quad Batch size per GPU (BW / Log / Lab / Sok) & 6 / 4 / 6 / 1 \\
\midrule
\multicolumn{2}{l}{\textit{Training}} \\
\quad Max sequence length & 14000 \\
\quad Epochs (BW / Log / Lab / Sok) & 100 / 100 / 100 / 100 \\
\bottomrule
\end{tabular}
\caption{Training hyperparameters for the generative model. Unless otherwise indicated, values are identical across domains (BW: Blocksworld, Log: Logistics, Lab: Labyrinth, Sok: Sokoban).}
\label{tab:pre_hyperparameters}
\end{table}

\begin{table}[h]
\centering
\begin{tabular}{lc}
\toprule
\textbf{Hyperparameter} & \textbf{Value} \\
\midrule
\multicolumn{2}{l}{\textit{Architecture}} \\
\quad Type & Transformer Encoder (BERT)  \\
\quad Transformer blocks ($n_\text{layer}$) & 6 \\
\quad Attention heads ($n_\text{head}$) & 12 \\
\quad Embedding dimension ($n_\text{embd}$) & 768 \\
\quad Feedforward dimension ($n_\text{inner}$) & 1536 \\
\midrule
\multicolumn{2}{l}{\textit{Optimization}} \\
\quad Learning rate & 5e-5 \\
\quad Scheduler & No \\
\quad Warmup steps & 1000 \\
\quad Optimizer & AdamW \\
\quad Batch size per GPU & 16 \\
\midrule
\multicolumn{2}{l}{\textit{Training}} \\
\quad Max sequence length & 14000 \\
\quad Epochs & 500 \\
\bottomrule
\end{tabular}
\caption{Training hyperparameters for the heuristic model. Unless otherwise indicated, values are identical across domains (BW: Blocksworld, Log: Logistics, Lab: Labyrinth, Sok: Sokoban).}
\label{tab:pre_hyperparameters_heuristic_model}
\end{table}

\clearpage
\section{Test-time Inference Hyperparameters}\label{app:evaluation_hyperparameters}
For all methods that employ rollouts, we use softmax temperature $T_\text{softmax}=1$ and sample plans in batches of 10. To further reduce the branching factor for highly complex problem instances, we use a simple heuristic based on the number of grounded actions in a problem instance. Specifically, if the number of grounded actions exceeds a predefined threshold $\tau_{|\mathcal{A}|}$ (in practice, this occurs only for a small portion of problems in Logistics), we perform expansions in \textsc{OCLGen} by sampling a batch of actions from $\pi_\theta$ with batch size 32; otherwise, we enumerate all successor states using the grounded operator models.

\begin{table}[h]
\centering
\begin{tabular}{lc}
\toprule
\textbf{Hyperparameter} & \textbf{Value} \\
\midrule
$N_\text{init}$ & 3 \\ 
Max tokens per rollout (BW/Log/Lab/Sok)  & 50 / 2000 / 50 / 2000 \\
Confidence threshold $\tau_\text{conf}$ (BW/Log/Lab/Sok) & 0.95 / 0.2 / 0.95 / 0.95  \\
Heuristic Percentile $k$ (BW/Log/Lab/Sok) & 3 / 10 / 3 / 10 \\ 
Policy expansion threshold $\tau_{|\mathcal{A}|}$ & $10^7$ \\
\bottomrule
\end{tabular}
\caption{Test-time hyperparameters for our method. Unless otherwise indicated, values are identical across domains (BW: Blocksworld, Log: Logistics, Lab: Labyrinth, Sok: Sokoban).}
\label{tab:test_time_hyperparameters}
\end{table}

\clearpage
\section{Self-improvement Details}\label{app:self_improvement_hyperparameters}

Our self-improvement pipeline runs for $n_{\text{loop}}=3$ iterations on Blocksworld and Sokoban. Every iteration samples 3000 problems per domain, executing the search method with $t_{\text{max}}=3$ minutes (Blocksworld) and $t_{\text{max}}=5$ minutes (Sokoban). We collect the successfully optimized plans to finetune both the generative model and the heuristic using standard cross-entropy loss. To maintain a consistent training set size, instances unsolved within the time limit default to their original suboptimal training plans. The respective training configurations are detailed in Tables~\ref{tab:ft_hyperparameters} and~\ref{tab:ft_hyperparameters_heuristic_model}.

\begin{table}[h]
\centering
\begin{tabular}{lc}
\toprule
\textbf{Hyperparameter} & \textbf{Value} \\
\midrule
\multicolumn{2}{l}{\textit{Optimization}} \\
\quad Learning rate & 1e-5 \\
\quad Optimizer & AdamW \\
\quad Batch size per GPU (BW / Sok) & 6 / 1 \\
\midrule
\multicolumn{2}{l}{\textit{Training}} \\
\quad Epochs (BW/ Sok) & 100 / 50 \\
\quad Finetuning dataset size & 3000 \\
\bottomrule
\end{tabular}
\caption{Finetuning hyperparameters for the generative model. Unless otherwise indicated, values are identical across domains.}
\label{tab:ft_hyperparameters}
\end{table}

\begin{table}[h]
\centering
\begin{tabular}{lc}
\toprule
\textbf{Hyperparameter} & \textbf{Value} \\
\midrule
\multicolumn{2}{l}{\textit{Optimization}} \\
\quad Learning rate & 1e-5 \\
\quad Optimizer & AdamW \\
\quad Batch size per GPU & 16 / 16 \\
\midrule
\multicolumn{2}{l}{\textit{Training}} \\
\quad Epochs (BW/ Sok) & 100 / 50 \\
\quad Finetuning dataset size & 3000 \\
\bottomrule
\end{tabular}
\caption{Finetuning hyperparameters for the heuristic model. Unless otherwise indicated, values are identical across domains.}
\label{tab:ft_hyperparameters_heuristic_model}
\end{table}

\clearpage
\section{Additional Results}

\subsection{Data Augmentation via Action Compilation} \label{app:action_compilation}

In Section~\ref{sec:base_model_improvements}, we introduce a data augmentation method for training our generative planning model. Given the plan sequences in $\mathcal{D}_\text{train}$, we use the operator effects specified in the PDDL domain definition to compile intermediate states along each plan, generating new state-plan pairs. We apply this procedure when training both the generative model and the heuristic model. During batch generation, we first sample a random offset within the plan length, then compute the corresponding intermediate state by sequentially applying operator effects up to this offset. A new training sample is formed by pairing this intermediate state with the remaining plan suffix.

The results in Table~\ref{tab:results_saw_vs_act} compare models trained with and without this state compilation augmentation. The evaluation, performed on 1000 unseen problem instances per domain, uses Best-of-N sampling with $N=10$ for both model variants. Outside of Blocksworld, where the two variants perform comparably, augmentation improves completion rate on Logistics, Labyrinth, and Sokoban, and shortens plans on Labyrinth and Sokoban. Overall, exposing the model to intermediate states during training enhances its ability to generalize across planning instances.

\begin{table}[h]
\centering
\resizebox{\textwidth}{!}{%
\begin{tabular}{l|cc|cc|cc|cc}
\toprule
& \multicolumn{2}{c|}{\textbf{Blocksworld}} & \multicolumn{2}{c|}{\textbf{Logistics}} & \multicolumn{2}{c|}{\textbf{Labyrinth}} & \multicolumn{2}{c}{\textbf{Sokoban}} \\
\textbf{GM Training Treatment} & Comp.[$\%$]  & Length & Comp. [$\%$] & Length & Comp. [$\%$] & Length & Comp. [$\%$] & Length \\
\midrule
w/o compilation & 100.0 & 69.06 (± 1.25) & 99.0 & 161.83 (± 3.60) & 98.60 & 25.44 (± 0.55) & 91.2 & 141.02 (± 3.58) \\
w/ compilation & 99.9 & 69.51 (± 1.24) & 99.8 & 162.17 (± 3.58)  & 99.90 & 24.23 (± 0.55) & 95.8 & 135.00 (± 3.38) \\
\bottomrule
\end{tabular}%
}
\caption{Performance comparison across problem domains for a generative planning model with and without action compilation. Comp.: Completion (\%); Length: Plan length (mean $\pm$ std error).}
\label{tab:results_saw_vs_act}
\end{table}

\subsection{Accuracy of the learned cost model}

\begin{figure}[H]
    \centering
    \includegraphics[width=0.5\linewidth]{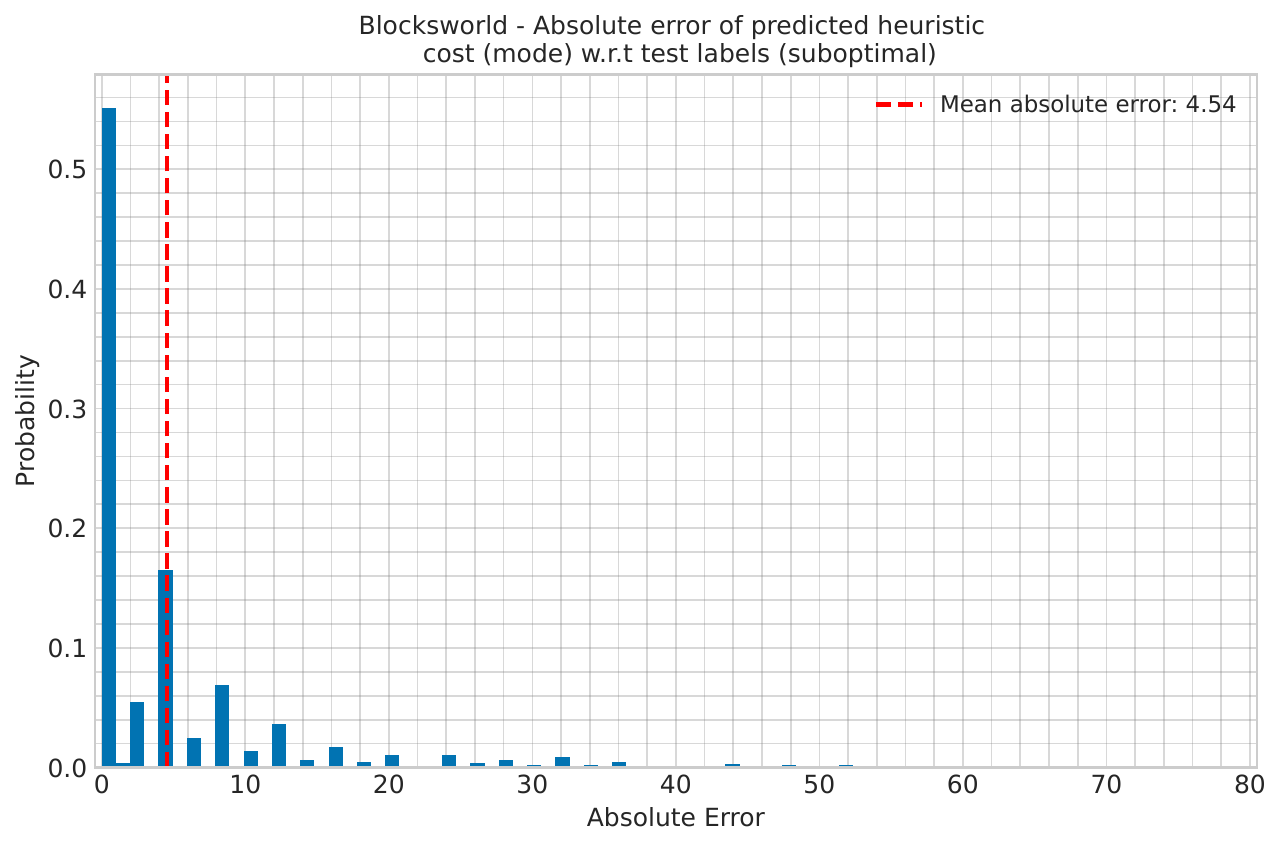}
    \caption{Blocksworld - Distribution of absolute errors of learned heuristic \\ cost model with respect to test labels (suboptimal data).}
    \label{fig:heuristic_cost_error_dist_blocksworld}
\end{figure}

\begin{figure}[H]
    \centering
    \includegraphics[width=0.5\linewidth]{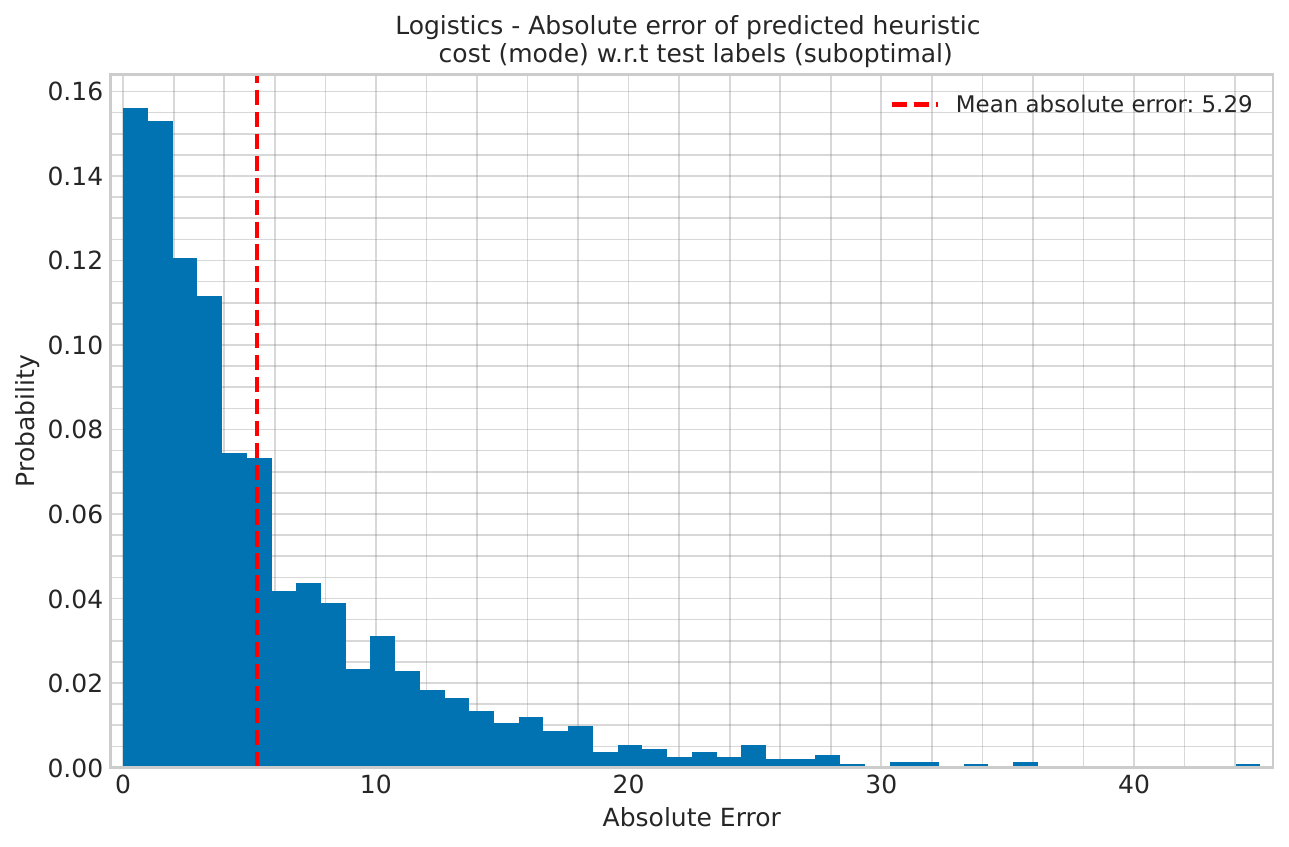}
    \caption{Logistics - Distribution of absolute errors of learned heuristic \\ cost model with respect to test labels (suboptimal data).}
    \label{fig:heuristic_cost_error_dist_logistics}
\end{figure}

\begin{figure}[H]
    \centering
    \includegraphics[width=0.5\linewidth]{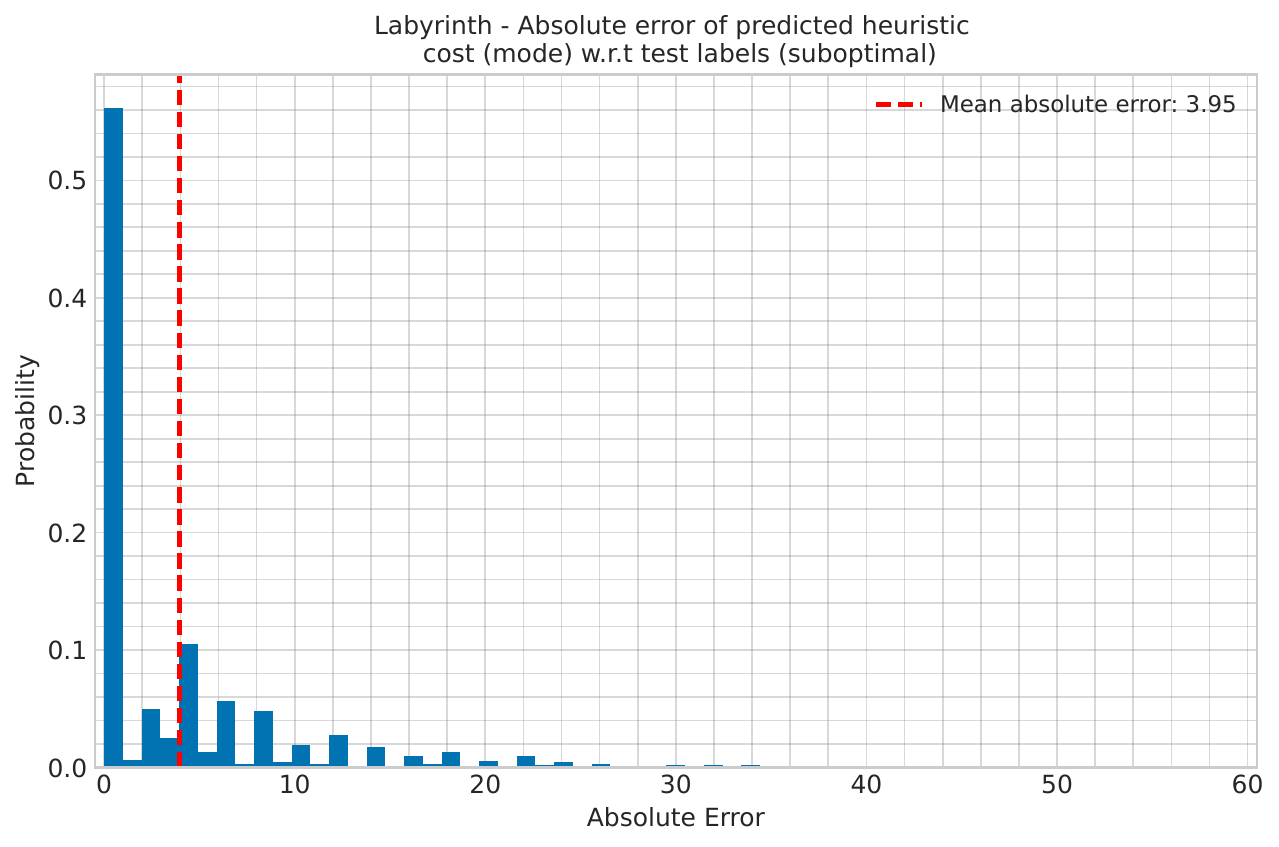}
    \caption{Labyrinth - Distribution of absolute errors of learned heuristic \\ cost model with respect to test labels (suboptimal data).}
    \label{fig:heuristic_cost_error_dist_labyrinth}
\end{figure}

\begin{figure}[H]
    \centering
    \includegraphics[width=0.5\linewidth]{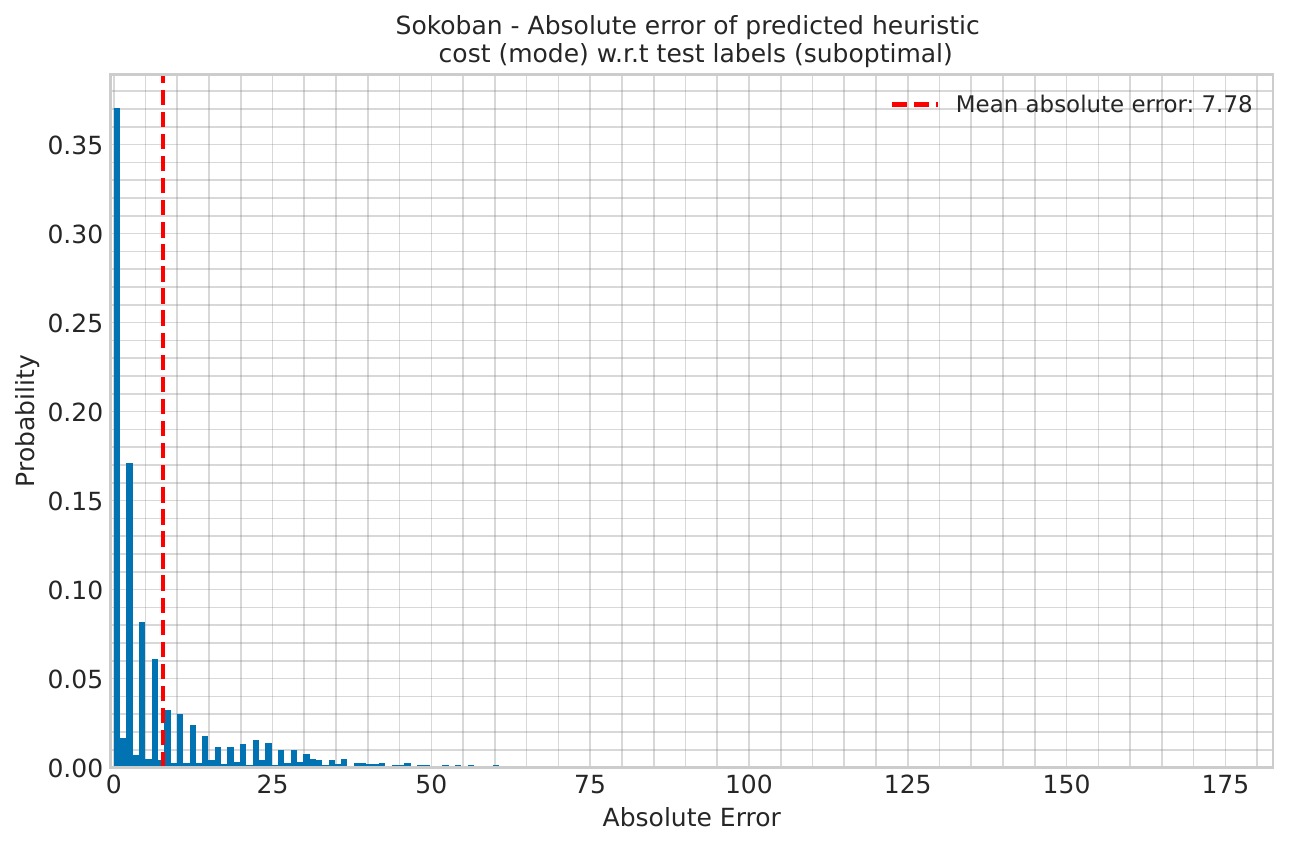}
    \caption{Sokoban - Distribution of absolute errors of learned heuristic \\ cost model with respect to test labels (suboptimal data).}
    \label{fig:heuristic_cost_error_dist_sokoban}
\end{figure}

Figures~\ref{fig:heuristic_cost_error_dist_blocksworld}--\ref{fig:heuristic_cost_error_dist_sokoban} show the per-domain distribution of absolute errors between $h_\phi$'s predictions and the held-out (suboptimal) test labels, characterizing how closely the heuristic fits the data-generating planner. Accuracy against the \emph{true optimal} cost-to-go is reported separately in Table~\ref{tab:mae_me_comparison}.

\clearpage
\subsection{Illustrations of Learned Cost Distribution}\label{app:cost_distributions}

\begin{figure}[H]
    \centering
    \includegraphics[width=0.6\linewidth]{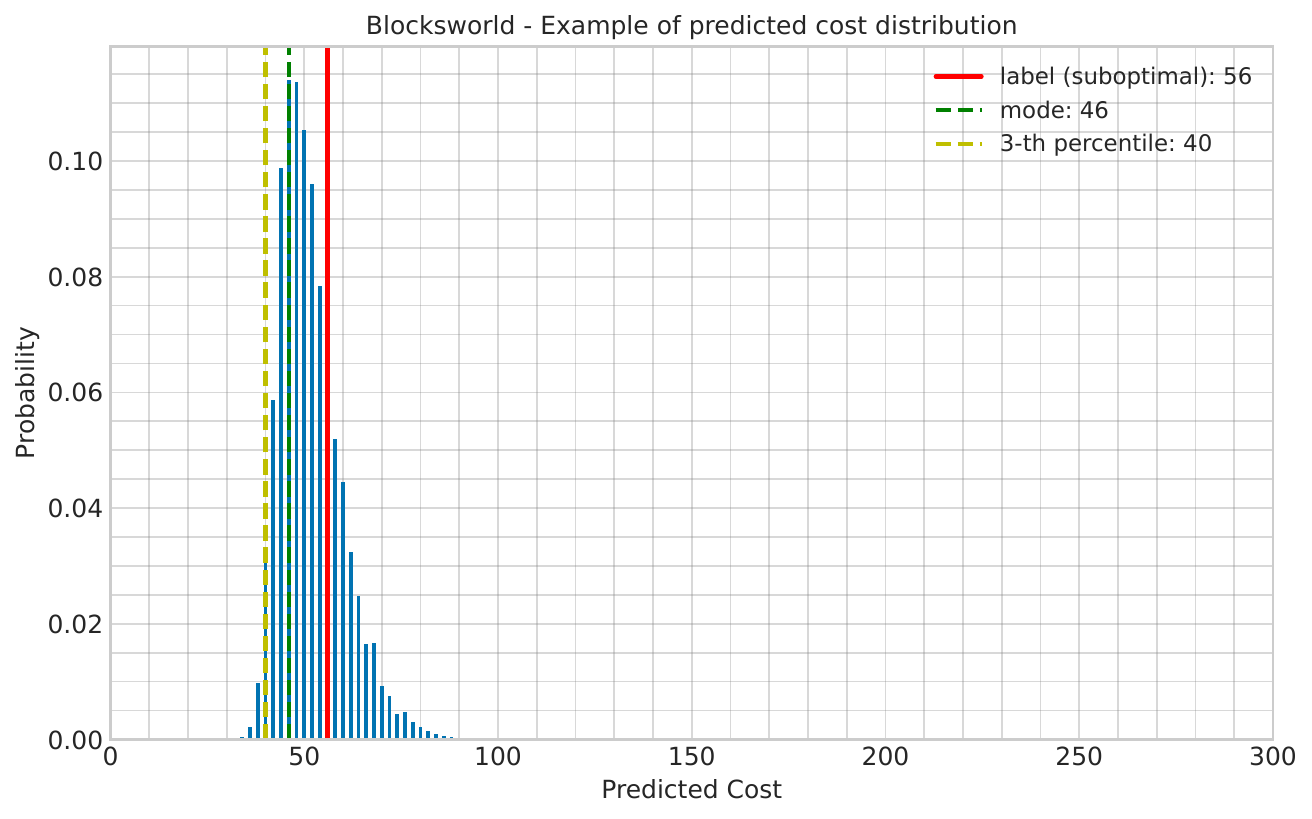}
    \includegraphics[width=0.6\linewidth]{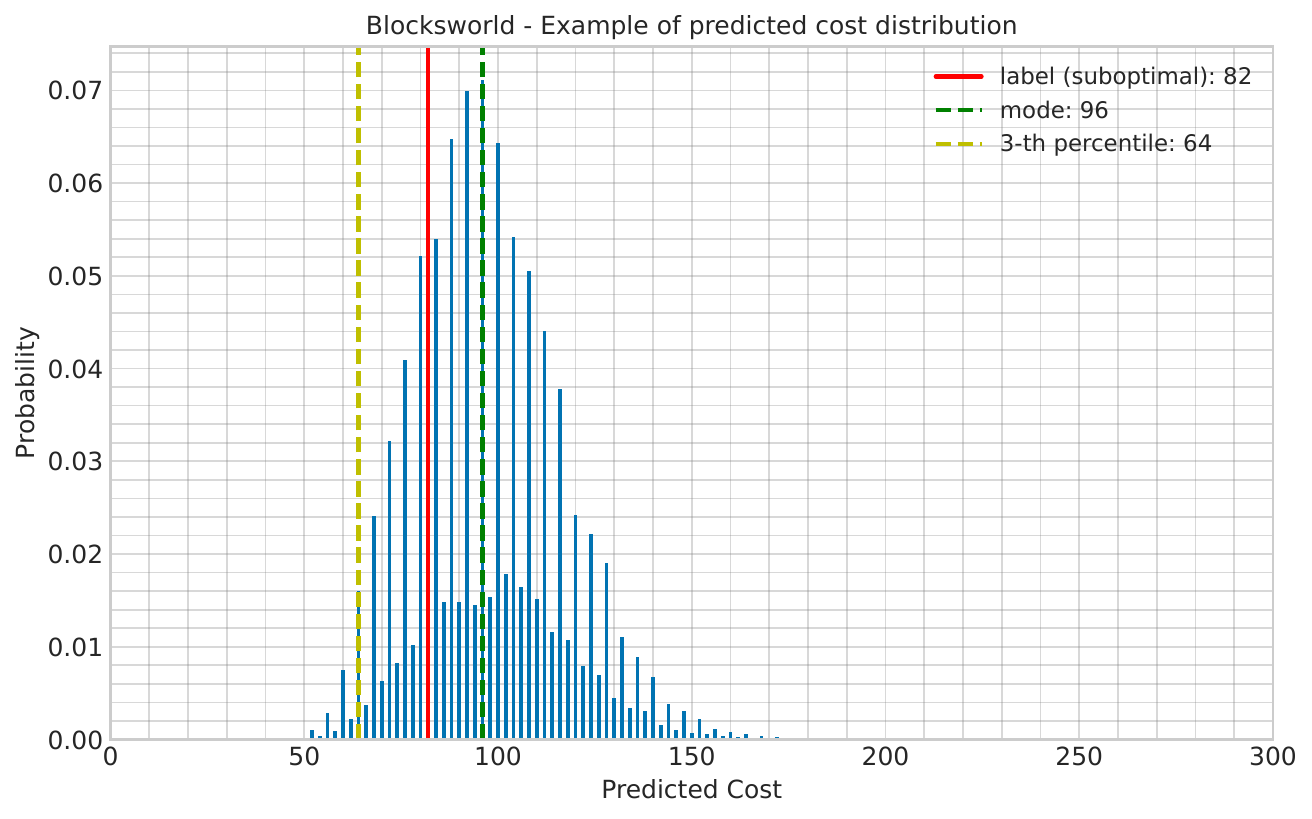}
    \includegraphics[width=0.6\linewidth]{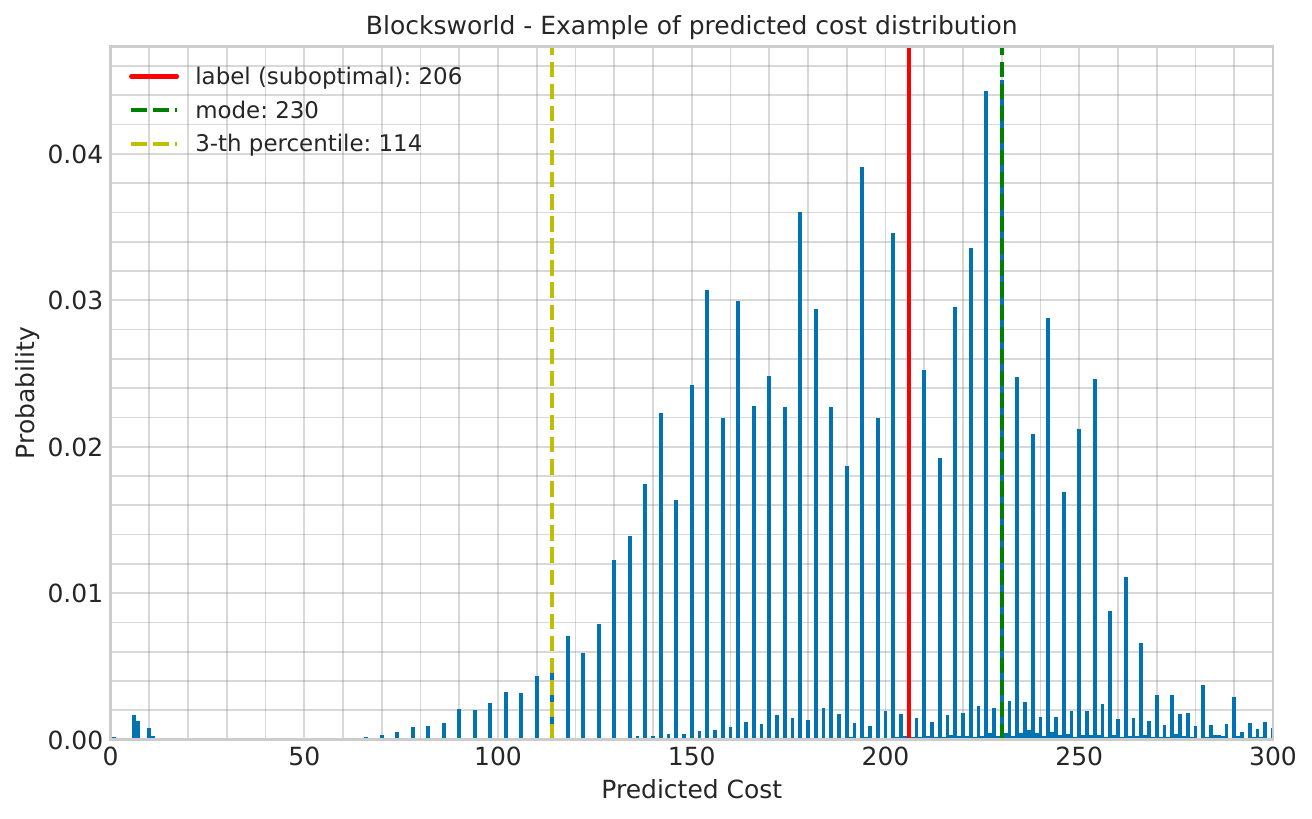}
    \caption{Blocksworld - Examples of cost distribution generated by the learned heuristic model.}
    \label{fig:heuristic_cost_dist_blocksworld}
\end{figure}

\begin{figure}[H]
    \centering
    \includegraphics[width=0.6\linewidth]{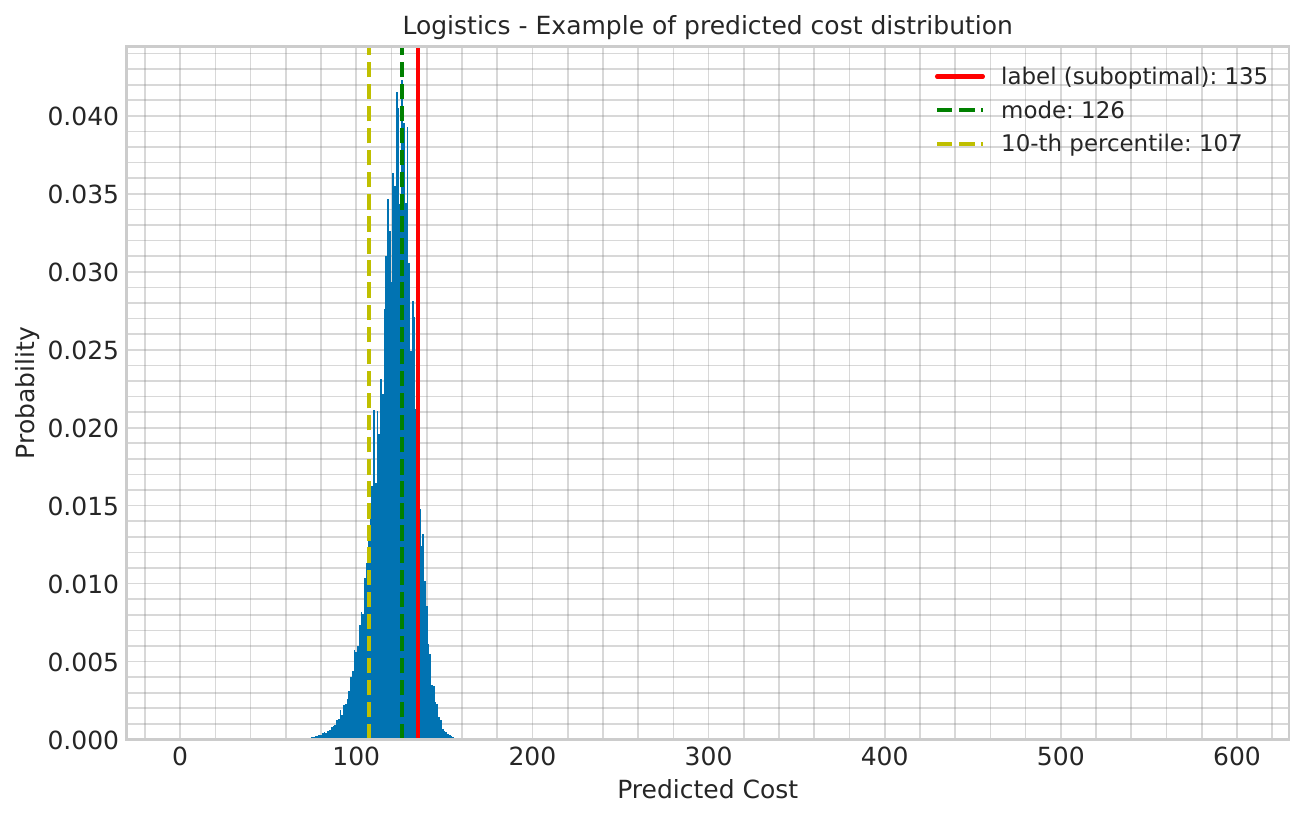}
    \includegraphics[width=0.6\linewidth]{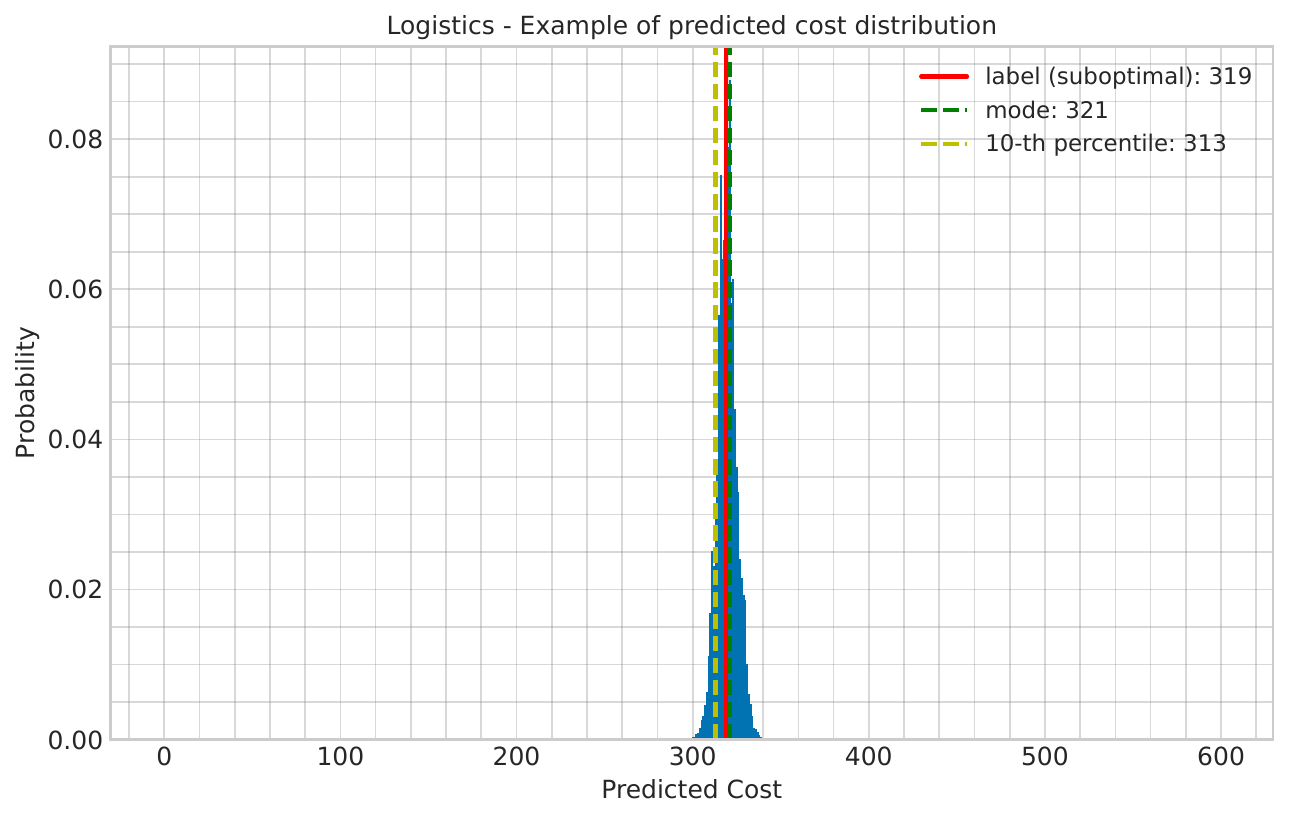}
    \includegraphics[width=0.6\linewidth]{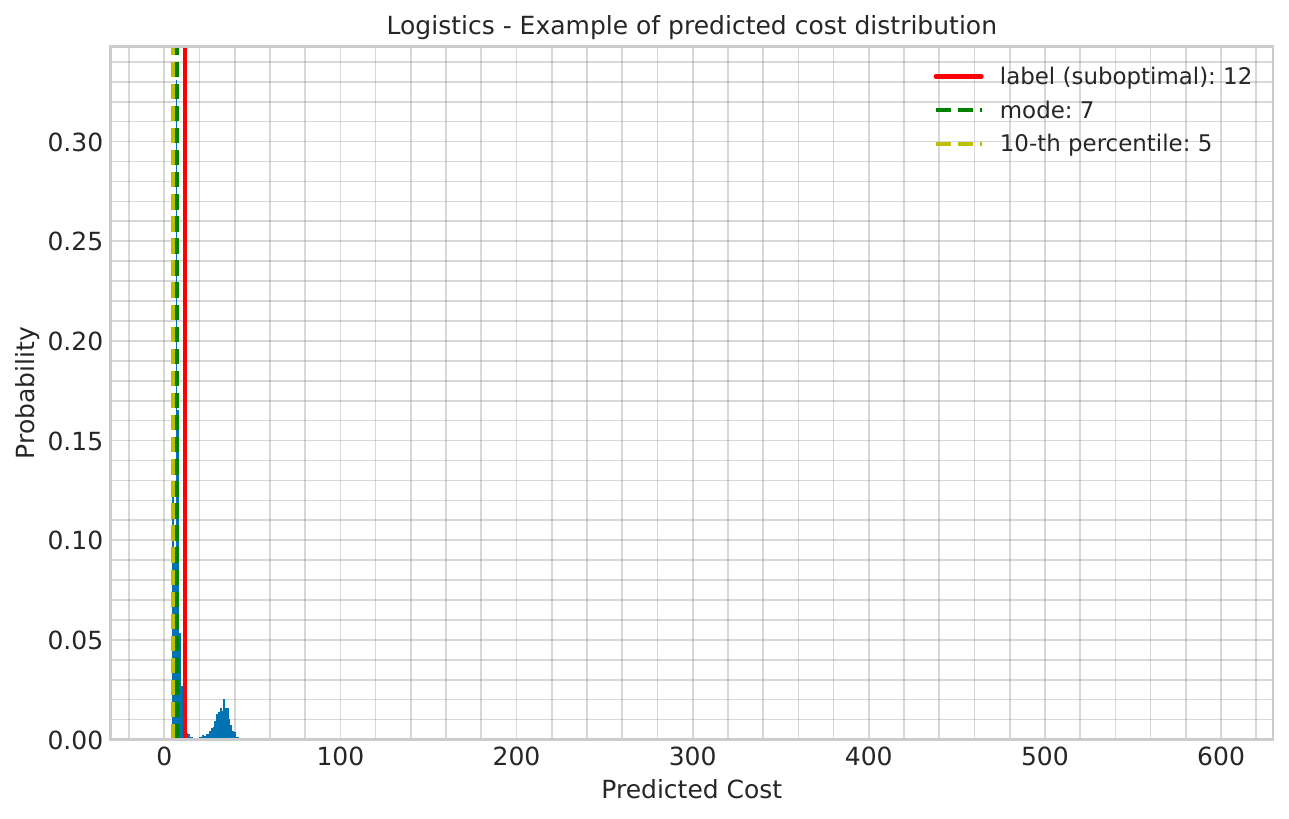}
    \caption{Logistics - Examples of cost distribution generated by the learned heuristic model.}
    \label{fig:heuristic_cost_dist_logistics}
\end{figure}

\begin{figure}[H]
    \centering
    \includegraphics[width=0.6\linewidth]{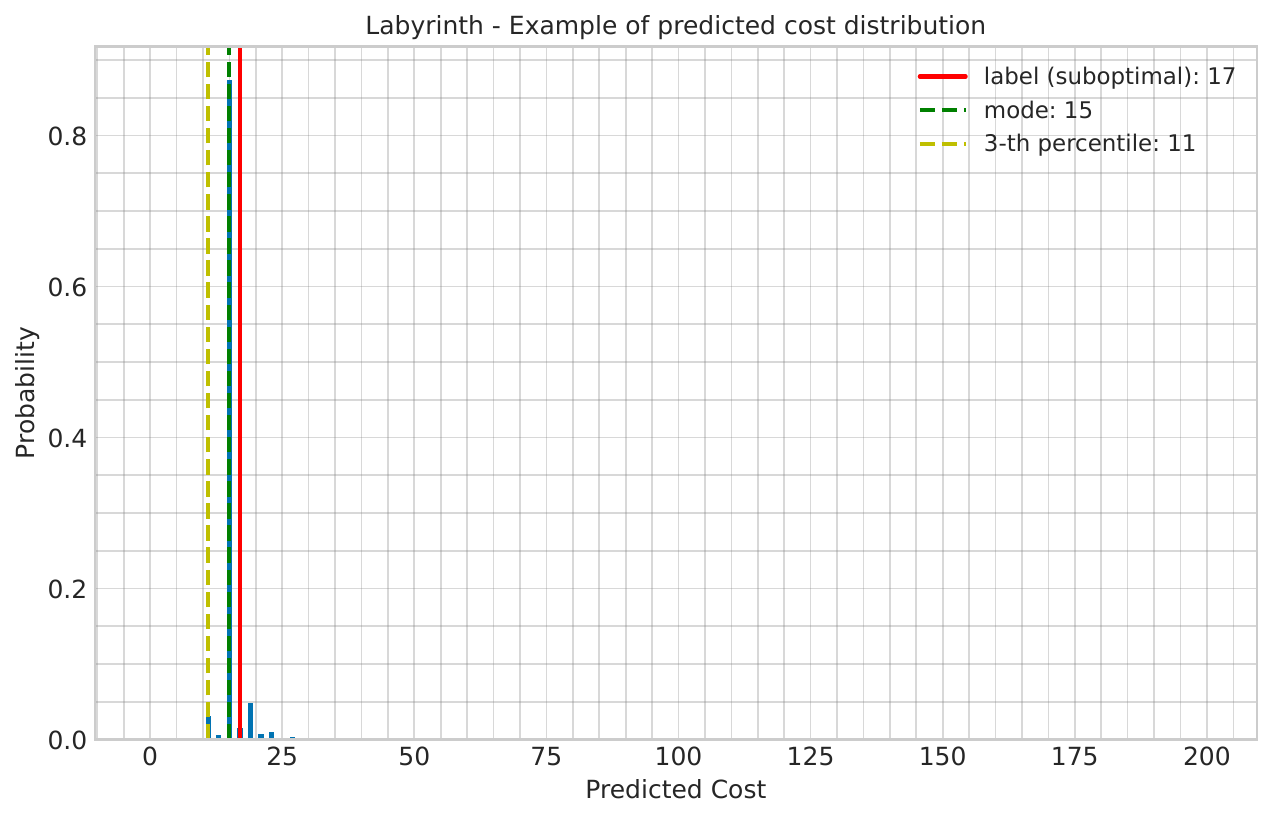}
    \includegraphics[width=0.6\linewidth]{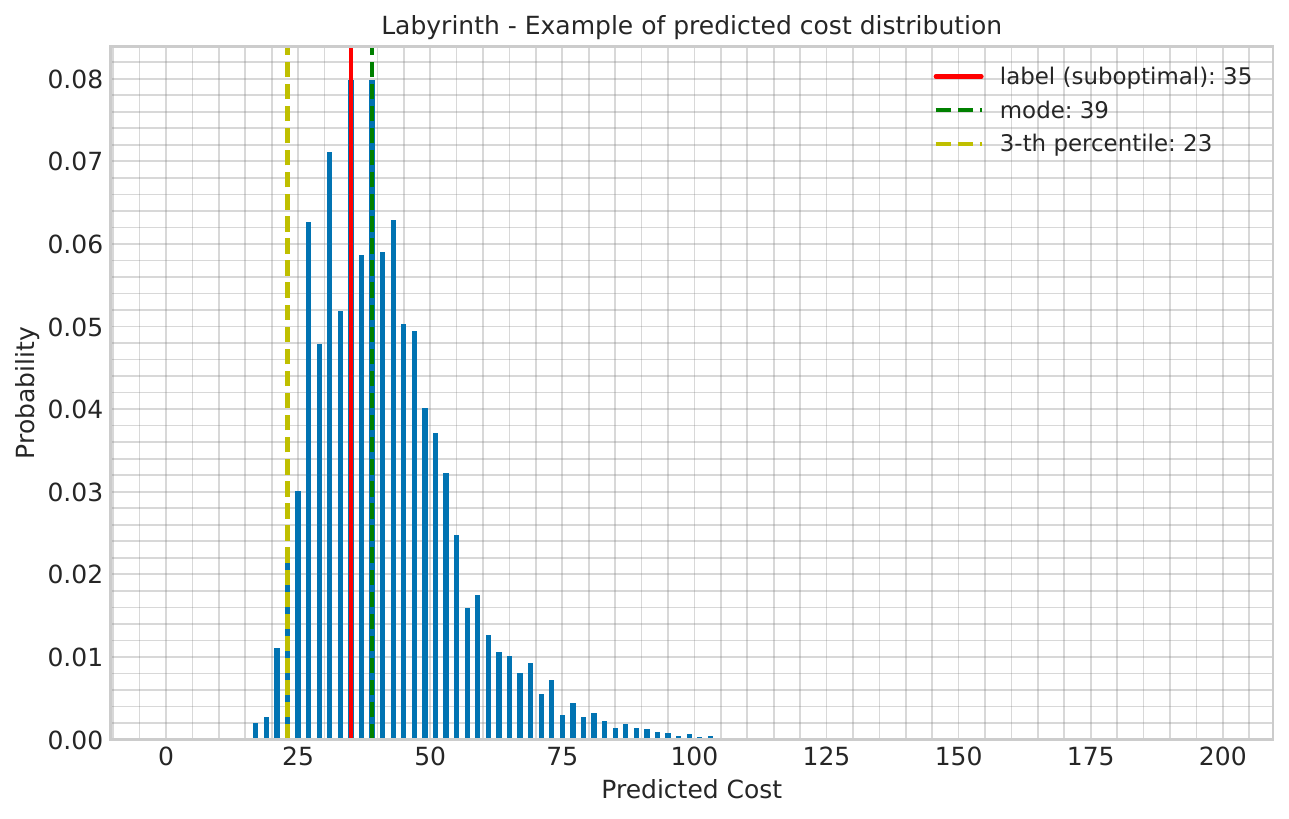}
    \includegraphics[width=0.6\linewidth]{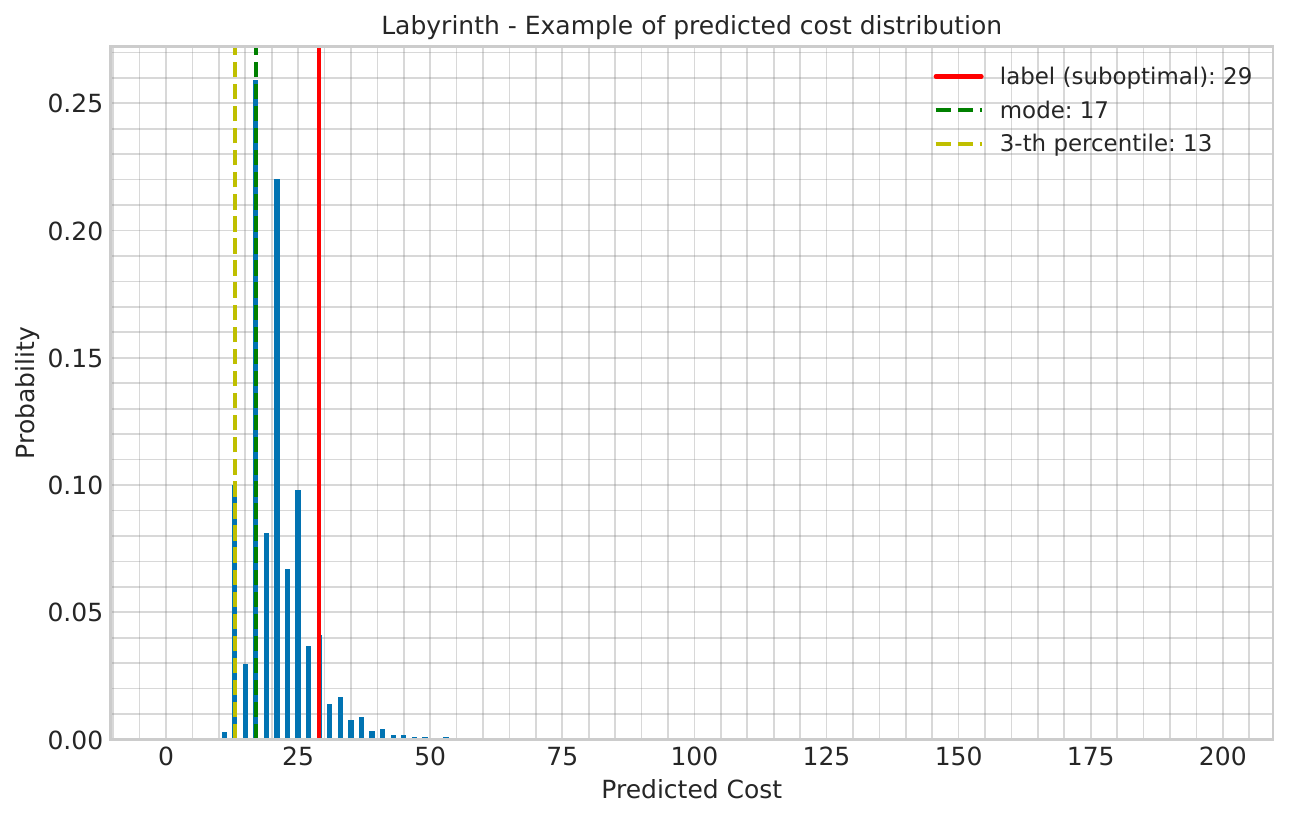}
    \caption{Labyrinth - Examples of cost distribution generated by the learned heuristic model.}
    \label{fig:heuristic_cost_dist_labyrinth}
\end{figure}

\begin{figure}[H]
    \centering
    \includegraphics[width=0.6\linewidth]{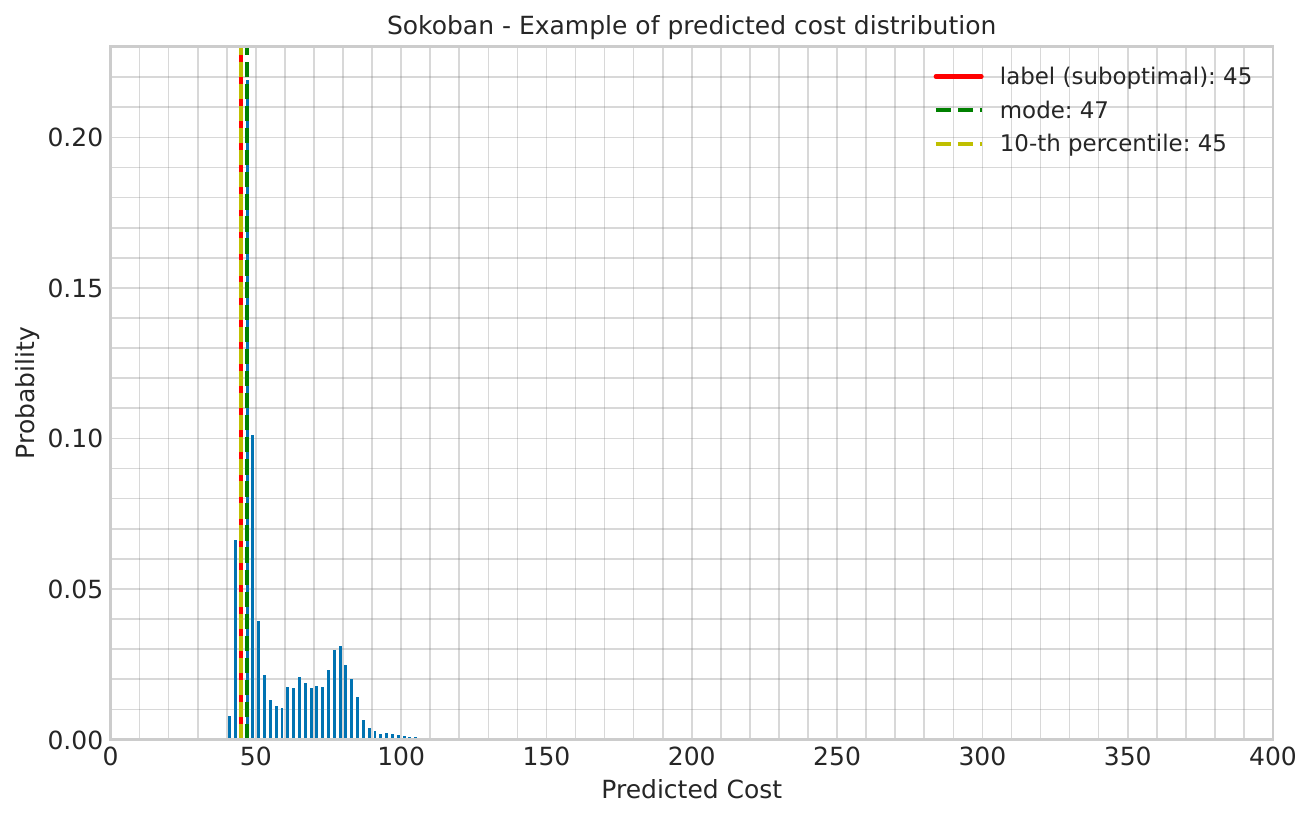}
    \includegraphics[width=0.6\linewidth]{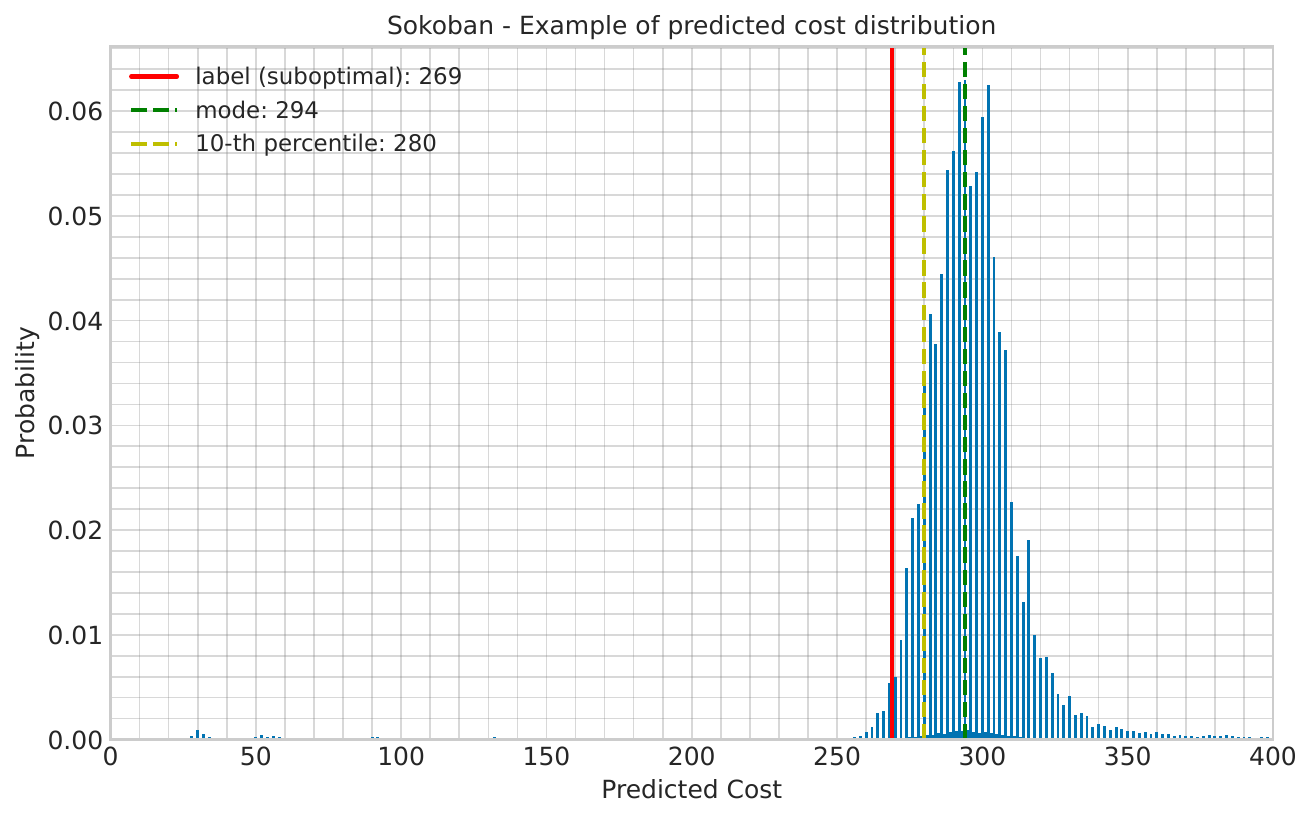}
    \includegraphics[width=0.6\linewidth]{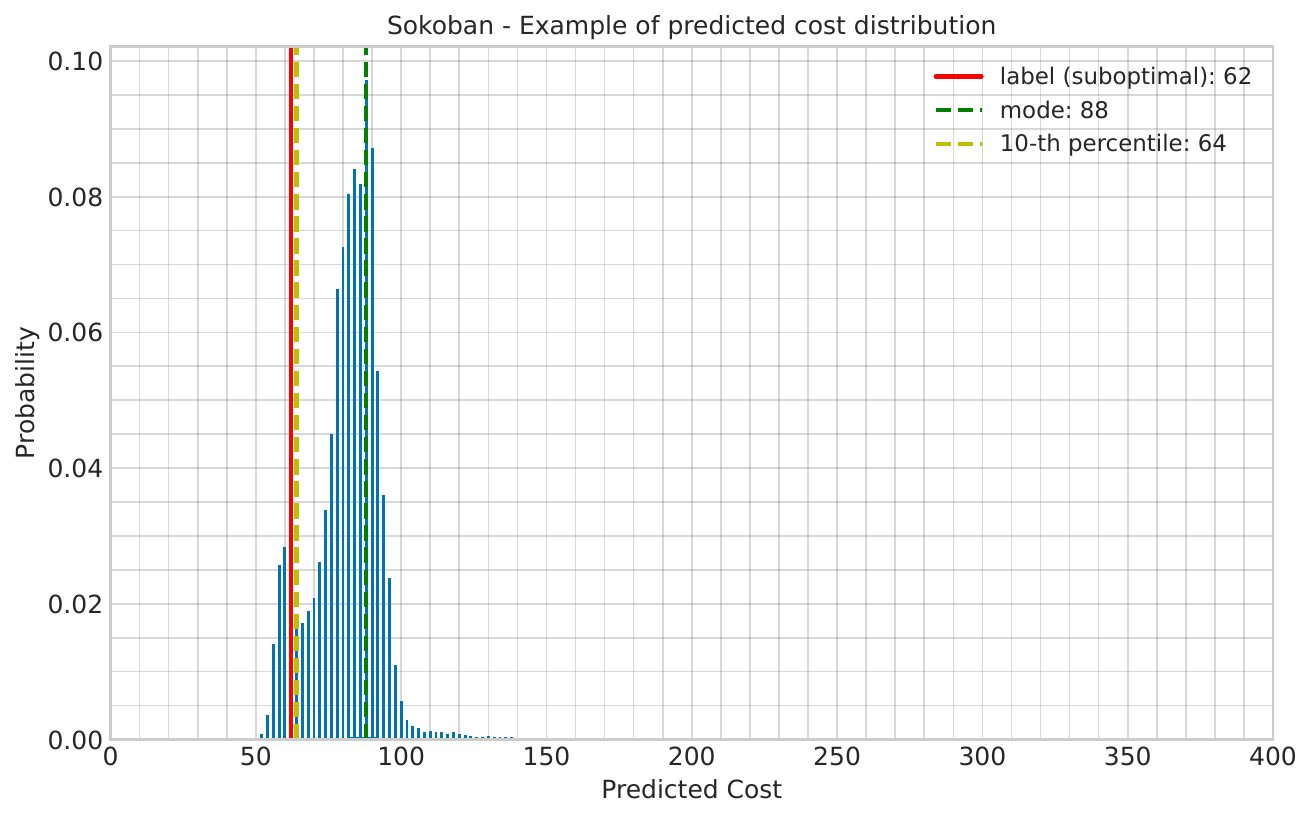}
    \caption{Sokoban - Examples of cost distribution generated by the learned heuristic model.}
    \label{fig:heuristic_cost_dist_sokoban}
\end{figure}

\clearpage
\subsection{Heuristic model vs. true optimal cost-to-go}

We compare the predictions of our learned heuristic model against the true optimal cost-to-go for the subset of solved problems with known optimal solutions. Table~\ref{tab:mae_me_comparison} reports the mean absolute error (MAE) and mean error (ME; $\text{prediction} - \text{ground truth}$) statistics across all domains. We evaluate both the predicted heuristic value with the highest probability (Mode) and the value obtained through percentile estimation (Perc.).

\begin{table}[h]
  \centering
  \begin{tabular}{llcccccc}
    \toprule
    &    & BW & BW ($n_\text{loop}{=}3$) & Log & Lab & Sok & Sok ($n_\text{loop}{=}3$) \\
    \midrule
    \multirow{2}{*}{MAE} 
     & Mode & $8.68 \pm 0.24$ & $0.44 \pm 0.02$ & $3.41 \pm 0.09$ & $2.32 \pm 0.10$ & $5.28 \pm 0.17$ & $3.05 \pm 0.11$ \\
     & Perc. & $4.07 \pm 0.12$ & $0.66 \pm 0.02$ & $3.07 \pm 0.08$ & $1.07 \pm 0.04$ & $3.31 \pm 0.12$ & $2.51 \pm 0.08$ \\
    \midrule
    \multirow{2}{*}{ME}
     & Mode & $8.68 \pm 0.24$ & $0.21 \pm 0.02$ & $1.40 \pm 0.11$ & $2.11 \pm 0.10$ & $4.79 \pm 0.17$ & $1.81 \pm 0.12$ \\
     & Perc. & $3.99 \pm 0.12$ & $-0.51 \pm 0.02$ & $-0.56 \pm 0.10$ & $0.56 \pm 0.05$ & $2.23 \pm 0.13$ & $-0.48 \pm 0.09$ \\
    \bottomrule
  \end{tabular}
  \caption{MAE (mean absolute error) and ME (mean error) statistics ($\pm$ std. error) with respect to the subset of known optimal data. Shown for both mode and percentile-based heuristic estimation across domains.}\label{tab:mae_me_comparison}
\end{table}

These results confirm the benefit of our percentile-based approach, which consistently shifts the predictions of the base heuristic model closer to the true optimal costs. Moreover, the self-improvement loop ($n_{\text{loop}}=3$) further reduces the estimation error of both configurations, substantially improving overall heuristic accuracy.

\subsection{Model Generalization to Larger Number of Objects}

Sec.~\ref{sec:generalization_lama_failed} shows that \textsc{OCLGen} generates valid plans where FD-LAMA failed, and outperforms Best-of-N sampling. Generalization beyond trained object counts is limited by the generative model's token vocabulary, which is a limitation shared by all baselines, not specific to our inference method.

Yet, we ran an additional experiment: training on Blocksworld with 3–25 blocks while excluding problems with 11, 12, 18, or 19 blocks, then testing on those (interpolate) and testing on instances with 26 to 40 blocks (extrapolate). The model with Best-of-N solved 100\% of interpolation instances but 0\% of extrapolation (unseen object names). The results on extrapolation are not surprising since the inputs during testing contain unobserved object name tokens (e.g. block31). In order to improve generalization to unseen object names, we retrained the base model while augmenting the data by randomly shuffling object names during the batch generation. With this new generative model, Best-of-N achieved 74.0\% completion on the extrapolation set within a 10min time limit per problem. Test-time search with \textsc{OCLGen} further improved the completion rate to 79.9\%.

\subsection{Statistical Analysis}
Table~\ref{tab:main_results} in Section~\ref{sec:experiments} presents the mean plan length and standard errors of all methods across different domains. Yet, in order to determine statistical significance, simply analyzing these statistics alone is insufficient since we test on the same 1000 instances per domain where problem difficulty dominates the variance. For example, the standard deviation of plan length is close to 110 on the Logistics test set. This shared structure means measurements are highly correlated across methods ($\rho > 0.81$, often $> 0.99$), so unpaired comparisons vastly inflate the standard error and mask consistent improvements. For example, in Logistics the unpaired $SE$ of the difference is $4.94$, while the paired $SE$ is just $0.11$ — a $45\times$ reduction. 

We therefore perform Wilcoxon signed-rank tests on paired per-problem differences $\Delta_i = L_{\text{OCLGen},i} - L_{\text{baseline},i}$, comparing OCLGen against MCTS (full rollouts), the strongest baseline with comparable completion rates:

\begin{table}[h]
  \centering
  \begin{tabular}{lcccc}
    \toprule
    Domain & $N$ & Mean $\Delta$ & Cohen's $d$ & Wilcoxon $p$ \\
    \midrule
    Blocksworld & 1000 & $-10.68 \pm 0.31$ & $-1.08$ & $5.4 \cdot 10^{-136}$ \\
    Logistics   & 1000 & $-2.92 \pm 0.11$  & $-0.81$ & $1.4 \cdot 10^{-110}$ \\
    Labyrinth   & 1000 & $-2.82 \pm 0.15$  & $-0.60$ & $1.0 \cdot 10^{-66}$  \\
    Sokoban     &  998 & $-2.55 \pm 0.21$  & $-0.39$ & $3.2 \cdot 10^{-48}$  \\
    \bottomrule
  \end{tabular}
  \caption{Statistical comparison of \textsc{OCLGen} vs.\ MCTS across domains.}
  \label{tab:statistical_comparison}
\end{table}

Table~\ref{tab:statistical_comparison_iter3} extends this comparison to a self-improvement setting, where both methods undergo 3 iterations of self-improvement using their respective internal search strategies (OCLGen vs. MCTS). 

\begin{table}[h]
  \centering
  \begin{tabular}{lcccc}
    \toprule
    Domain & $N$ & Mean $\Delta$ & Cohen's $d$ & Wilcoxon $p$ \\
    \midrule
    Blocksworld & 1000 & $-4.15 \pm 0.19$ & $-0.69$ & $4.1 \times 10^{-89}$ \\
    Sokoban     &  992 & $-2.08 \pm 0.13$ & $-0.51$ & $2.6 \times 10^{-56}$ \\
    \bottomrule
  \end{tabular}
  \caption{Statistical comparison of \textsc{OCLGen} vs.\ MCTS (self-improvement, 3 iterations) across domains.}\label{tab:statistical_comparison_iter3}
\end{table}

Across all domains and both experimental settings, the Wilcoxon signed-rank test returns p-values that are vanishingly small, ranging from $5.4 \times 10^{-136}$ in Blocksworld to $3.2 \times 10^{-48}$ in Sokoban (Table~\ref{tab:statistical_comparison}). All results comfortably exceed any conventional significance threshold (e.g., $\alpha = 0.05$ or even $\alpha = 0.001$). The self-improvement setting (Table~\ref{tab:statistical_comparison_iter3}) confirms this pattern, with p-values of $4.1 \times 10^{-89}$ and $2.6 \times 10^{-56}$ for Blocksworld and Sokoban respectively.

The Cohen's $d$ values provide a far more informative picture of OCLGen's advantage. Using the conventional benchmarks ($|d| = 0.2$ small, $|d| = 0.5$ medium, $|d| = 0.8$ large), the results reveal a consistent but domain-dependent advantage for OCLGen.

In Table~\ref{tab:statistical_comparison}, Blocksworld shows the strongest effect ($d = -1.08$), a large effect by any standard, accompanied by the largest absolute mean difference of $-10.68 \pm 0.31$ steps. Logistics follows with a large effect ($d = -0.81$, $\Delta = -2.92 \pm 0.11$), reflecting a similar advantage. Labyrinth yields a medium-to-large effect ($d = -0.60$, $\Delta = -2.82 \pm 0.15$), while Sokoban shows a smaller effect ($d = -0.39$, $\Delta = -2.55 \pm 0.21$), falling in the small-to-medium range.

In Table~\ref{tab:statistical_comparison_iter3}, both methods undergo 3 iterations of refinement using their own internal strategy before being compared head-to-head. OCLGen maintains a meaningful advantage over self-improved MCTS in both tested domains. Blocksworld yields $d = -0.69$ (medium-to-large) with $\Delta = -4.15 \pm 0.19$, and Sokoban yields $d = -0.51$ (medium) with $\Delta = -2.08 \pm 0.13$. 

Taken together, the results demonstrate that OCLGen consistently and substantially outperforms MCTS in plan length across all tested domains. The statistical evidence is unambiguous, and the effect sizes range from practically meaningful ($d \approx -0.39$ in Sokoban) to large ($d \approx -1.08$ in Blocksworld). When both methods are given equal opportunity to self-improve over 3 iterations, OCLGen retains its advantage with medium-to-large effects in both domains. 

\clearpage\section{PDDL Domain files} \label{app:domains}

\lstset{
  basicstyle=\ttfamily\footnotesize,
  breaklines=true,
  breakatwhitespace=true,
  columns=fullflexible,
  postbreak=\mbox{\textcolor{gray}{$\hookrightarrow$}\space},
  showstringspaces=false,
}

\subsection{Blocksworld}

\begin{lstlisting}[caption=Blocksworld domain file]
(define (domain blocksworld-4ops)
  (:requirements :strips)
(:predicates (clear ?x)
             (on-table ?x)
             (arm-empty)
             (holding ?x)
             (on ?x ?y))

(:action pickup
  :parameters (?ob)
  :precondition (and (clear ?ob) (on-table ?ob) (arm-empty))
  :effect (and (holding ?ob) (not (clear ?ob)) (not (on-table ?ob)) 
               (not (arm-empty))))

(:action putdown
  :parameters  (?ob)
  :precondition (holding ?ob)
  :effect (and (clear ?ob) (arm-empty) (on-table ?ob) 
               (not (holding ?ob))))

(:action stack
  :parameters  (?ob ?underob)
  :precondition (and (clear ?underob) (holding ?ob))
  :effect (and (arm-empty) (clear ?ob) (on ?ob ?underob)
               (not (clear ?underob)) (not (holding ?ob))))

(:action unstack
  :parameters  (?ob ?underob)
  :precondition (and (on ?ob ?underob) (clear ?ob) (arm-empty))
  :effect (and (holding ?ob) (clear ?underob)
               (not (on ?ob ?underob)) (not (clear ?ob)) (not (arm-empty)))))
\end{lstlisting}

\clearpage
\subsection{Logistics}

\begin{lstlisting}[caption=Logistics domain file]
(define (domain logistics)
  (:requirements :strips :typing)
  (:types 
  	package location vehicle - object
  	truck airplane - vehicle
  	city airport - location)
  
  (:predicates 	
		(at ?vehicle-or-package - (either vehicle package)  ?location - location)
		(in ?package - package ?vehicle - vehicle)
		(in-city ?loc-or-truck - (either location truck) ?citys - city))
		
  (:action load-truck
	:parameters
		 (?obj - package
		  ?truck - truck
		  ?loc - location)
	:precondition
		(and 	(at ?truck ?loc) 
			(at ?obj ?loc))
	:effect
		(and 	(not (at ?obj ?loc)) 
			(in ?obj ?truck)))

  (:action load-airplane
	:parameters
		(?obj - package
		 ?airplane - airplane
		 ?loc - airport)
	:precondition
		(and
			(at ?obj ?loc) 
			(at ?airplane ?loc))
	:effect
   		(and 	(not (at ?obj ?loc)) 
			(in ?obj ?airplane)))

  (:action unload-truck
	:parameters
		(?obj - package
		 ?truck - truck
		 ?loc - location)
	:precondition
		(and    (at ?truck ?loc) 
			(in ?obj ?truck))
	:effect
		(and	(not (in ?obj ?truck)) 
			(at ?obj ?loc)))

  (:action unload-airplane
	:parameters
		(?obj - package
		 ?airplane - airplane
		 ?loc - airport)
	:precondition
		(and	(in ?obj ?airplane) 
			(at ?airplane ?loc))
	:effect
		(and 
			(not (in ?obj ?airplane)) 
			(at ?obj ?loc)))

  (:action drive-truck
	:parameters
		(?truck - truck
		 ?loc-from - location
		 ?loc-to - location
		 ?city - city)
	:precondition
		(and 	(at ?truck ?loc-from)
			(in-city ?loc-from ?city)
			(in-city ?loc-to ?city))
	:effect
		(and 	(not (at ?truck ?loc-from)) 
			(at ?truck ?loc-to)))

  (:action fly-airplane
	:parameters
		(?airplane - airplane
		 ?loc-from - airport
		 ?loc-to - airport)
	:precondition
		(at ?airplane ?loc-from)
	:effect
		(and 	(not (at ?airplane ?loc-from)) 
		(at ?airplane ?loc-to)))
)
\end{lstlisting}

\clearpage
\subsection{Labyrinth}

\begin{lstlisting}[caption=Labyrinth domain file]
(define (domain labyrinth)
(:requirements :adl :action-costs :strips :typing :numeric-fluents)

(:types
    card - object
    direction - object
    directionV - direction
    directionH - direction
    gridpos - object
)

(:constants
    s n - directionV
    w e - directionH
)

(:predicates
    (next ?p1 - gridpos ?p2 - gridpos)
    (max-pos ?p - gridpos)
    (min-pos ?p - gridpos)
    (blocked ?c - card ?d - direction)
    (robot-at ?c - card)
    (card-at ?c - card ?x - gridpos ?y - gridpos)
    (left)
    (cards-moving)
    (cards-moving-west)
    (cards-moving-east)
    (cards-moving-south)
    (cards-moving-north)
    (next-moving-card ?c - card)
    (new-headtail-card ?c - card)

)

(:functions
    (total-cost) - number
    (move-robot-cost) - number
    (move-card) - number
)

(:action move-west
    :parameters (?cfrom - card ?xfrom - gridpos ?yfrom - gridpos ?dfrom - directionH ?cto - card ?xto - gridpos ?yto - gridpos ?dto - directionH)
    :precondition
        (and
            (not (cards-moving))
            (= ?dfrom w)
            (robot-at ?cfrom)
            (card-at ?cfrom ?xfrom ?yfrom)
            (card-at ?cto ?xto ?yto)
            (next ?xfrom ?xto)
            (= ?yfrom ?yto)
            (not (= ?dfrom ?dto))
            (not (blocked ?cfrom ?dfrom))
            (not (blocked ?cto ?dto))
        )
    :effect
        (and
            (not (robot-at ?cfrom))
            (robot-at ?cto)
            (increase (total-cost) (move-robot-cost))
        )
)

(:action move-east
    :parameters (?cfrom - card ?xfrom - gridpos ?yfrom - gridpos ?dfrom - directionH ?cto - card ?xto - gridpos ?yto - gridpos ?dto - directionH)
    :precondition
        (and
            (not (cards-moving))
            (= ?dfrom e)
            (robot-at ?cfrom)
            (card-at ?cfrom ?xfrom ?yfrom)
            (card-at ?cto ?xto ?yto)
            (next ?xto ?xfrom)
            (= ?yfrom ?yto)
            (not (= ?dfrom ?dto))
            (not (blocked ?cfrom ?dfrom))
            (not (blocked ?cto ?dto))
        )
    :effect
        (and
            (not (robot-at ?cfrom))
            (robot-at ?cto)
            (increase (total-cost) (move-robot-cost))
        )
)

(:action move-north
    :parameters (?cfrom - card ?xfrom - gridpos ?yfrom - gridpos ?dfrom - directionV ?cto - card ?xto - gridpos ?yto - gridpos ?dto - directionV)
    :precondition
        (and
            (not (cards-moving))
            (= ?dfrom n)
            (robot-at ?cfrom)
            (card-at ?cfrom ?xfrom ?yfrom)
            (card-at ?cto ?xto ?yto)
            (next ?yfrom ?yto)
            (= ?xfrom ?xto)
            (not (= ?dfrom ?dto))
            (not (blocked ?cfrom ?dfrom))
            (not (blocked ?cto ?dto))
        )
    :effect
        (and
            (not (robot-at ?cfrom))
            (robot-at ?cto)
            (increase (total-cost) (move-robot-cost))
        )
)

(:action move-south
    :parameters (?cfrom - card ?xfrom - gridpos ?yfrom - gridpos ?dfrom - directionV ?cto - card ?xto - gridpos ?yto - gridpos ?dto - directionV)
    :precondition
        (and
            (not (cards-moving))
            (= ?dfrom s)
            (robot-at ?cfrom)
            (card-at ?cfrom ?xfrom ?yfrom)
            (card-at ?cto ?xto ?yto)
            (next ?yto ?yfrom)
            (= ?xfrom ?xto)
            (not (= ?dfrom ?dto))
            (not (blocked ?cfrom ?dfrom))
            (not (blocked ?cto ?dto))
        )
    :effect
        (and
            (not (robot-at ?cfrom))
            (robot-at ?cto)
            (increase (total-cost) (move-robot-cost))
        )
)

(:action start-move-card-west
:parameters(?cm - card ?x - gridpos ?y - gridpos ?cnext - card ?nextx - gridpos)
:precondition
    (and
        (not (cards-moving))
        (not (cards-moving-west))
        (not (robot-at ?cm))
        (card-at ?cm ?x ?y )
        (min-pos ?x)
        (card-at ?cnext ?nextx ?y)
        (next ?nextx ?x)
    )
:effect
    (and
        (cards-moving)
        (cards-moving-west)
        (not (card-at ?cm ?x ?y ))
        (new-headtail-card ?cm)
        (next-moving-card ?cnext)
        (increase (total-cost) (move-card))
    )
)

(:action move-card-west
:parameters(?cm - card ?x - gridpos ?y - gridpos ?cnext - card ?nextx - gridpos ?prevx - gridpos)
:precondition
    (and
        (cards-moving)
        (cards-moving-west)
        (not (robot-at ?cm))
        (next-moving-card ?cm)
        (card-at ?cm ?x ?y )
        (card-at ?cnext ?nextx ?y)
        (next ?x ?prevx)
        (next ?nextx ?x)
    )
:effect
    (and
        (cards-moving)
        (cards-moving-west)
        (not (card-at ?cm ?x ?y))
        (card-at ?cm ?prevx ?y)
        (not (next-moving-card ?cm))
        (next-moving-card ?cnext)
        (increase (total-cost) (move-card))
    )
)

(:action stop-move-card-west
:parameters(?cm - card ?x - gridpos ?y - gridpos ?prevx - gridpos ?newtc - card)
:precondition
    (and
        (cards-moving)
        (cards-moving-west)
        (not (robot-at ?cm))
        (next-moving-card ?cm)
        (card-at ?cm ?x ?y )
        (next ?x ?prevx)
        (max-pos ?x)
        (new-headtail-card ?newtc)
    )
:effect
    (and
        (not (cards-moving))
        (not (cards-moving-west))
        (not (card-at ?cm ?x ?y))
        (card-at ?cm ?prevx ?y)
        (card-at ?newtc ?x ?y)
        (not (new-headtail-card ?newtc))
        (not (next-moving-card ?cm))
        (increase (total-cost) (move-card)) 
    )
)

(:action start-move-card-east
:parameters(?cm - card ?x - gridpos ?y - gridpos ?cnext - card ?nextx - gridpos)
:precondition
    (and
        (not (cards-moving))
        (not (cards-moving-east))
        (not (robot-at ?cm))
        (card-at ?cm ?x ?y )
        (max-pos ?x)
        (card-at ?cnext ?nextx ?y)
        (next ?x ?nextx)
    )
:effect
    (and
        (cards-moving)
        (cards-moving-east)
        (not (card-at ?cm ?x ?y ))
        (new-headtail-card ?cm)
        (next-moving-card ?cnext)
        (increase (total-cost) (move-card))
    )
)

(:action move-card-east
:parameters(?cm - card ?x - gridpos ?y - gridpos ?cnext - card ?nextx - gridpos ?prevx - gridpos)
:precondition
    (and
        (cards-moving)
        (cards-moving-east)
        (not (robot-at ?cm))
        (next-moving-card ?cm)
        (card-at ?cm ?x ?y )
        (card-at ?cnext ?nextx ?y)
        (next ?prevx ?x)
        (next ?x ?nextx)
    )
:effect
    (and
        (cards-moving)
        (cards-moving-east)
        (not (card-at ?cm ?x ?y))
        (card-at ?cm ?prevx ?y)
        (not (next-moving-card ?cm))
        (next-moving-card ?cnext)
        (increase (total-cost) (move-card))
    )
)

(:action stop-move-card-east
:parameters(?cm - card ?x - gridpos ?y - gridpos ?prevx - gridpos ?newtc - card)
:precondition
    (and
        (cards-moving)
        (cards-moving-east)
        (not (robot-at ?cm))
        (next-moving-card ?cm)
        (card-at ?cm ?x ?y )
        (next  ?prevx ?x)
        (min-pos ?x)
        (new-headtail-card ?newtc)
    )
:effect
    (and
        (not (cards-moving))
        (not (cards-moving-east))
        (not (card-at ?cm ?x ?y))
        (card-at ?cm ?prevx ?y)
        (card-at ?newtc ?x ?y)
        (not (new-headtail-card ?newtc))
        (not (next-moving-card ?cm))
        (increase (total-cost) (move-card))
    )
)

(:action start-move-card-north
:parameters(?cm - card ?x - gridpos ?y - gridpos ?cnext - card ?nexty - gridpos)
:precondition
    (and
        (not (cards-moving))
        (not (cards-moving-north))
        (not (robot-at ?cm))
        (card-at ?cm ?x ?y )
        (min-pos ?y)
        (card-at ?cnext ?x ?nexty)
        (next ?nexty ?y)
    )
:effect
    (and
        (cards-moving)
        (cards-moving-north)
        (not (card-at ?cm ?x ?y ))
        (new-headtail-card ?cm)
        (next-moving-card ?cnext)
        (increase (total-cost) (move-card))
    )
)

(:action move-card-north
:parameters(?cm - card ?x - gridpos ?y - gridpos  ?cnext - card ?nexty - gridpos ?prevy - gridpos)
:precondition
    (and
        (cards-moving)
        (cards-moving-north)
        (not (robot-at ?cm))
        (next-moving-card ?cm)
        (card-at ?cm ?x ?y )
        (card-at ?cnext ?x ?nexty)
        (next ?y ?prevy)
        (next ?nexty ?y)
    )
:effect
    (and
        (cards-moving)
        (cards-moving-north)
        (not (card-at ?cm ?x ?y))
        (card-at ?cm ?x ?prevy)
        (not (next-moving-card ?cm))
        (next-moving-card ?cnext)
        (increase (total-cost) (move-card))
    )
)

(:action stop-move-card-north
:parameters(?cm - card ?x - gridpos ?y - gridpos ?prevy - gridpos ?newtc - card)
:precondition
    (and
        (cards-moving)
        (cards-moving-north)
        (not (robot-at ?cm))
        (next-moving-card ?cm)
        (card-at ?cm ?x ?y )
        (next ?y ?prevy)
        (max-pos ?y)
        (new-headtail-card ?newtc)
    )
:effect
    (and
        (not (cards-moving))
        (not (cards-moving-north))
        (not (card-at ?cm ?x ?y))
        (card-at ?cm ?x ?prevy)
        (card-at ?newtc ?x ?y)
        (not (new-headtail-card ?newtc))
        (not (next-moving-card ?cm))
        (increase (total-cost) (move-card))
    )
)

(:action start-move-card-south
:parameters(?cm - card ?x - gridpos ?y - gridpos  ?cnext - card ?nexty - gridpos)
:precondition
    (and
        (not (cards-moving))
        (not (cards-moving-south))
        (not (robot-at ?cm))
        (card-at ?cm ?x ?y )
        (max-pos ?y)
        (card-at ?cnext ?x ?nexty)
        (next ?y ?nexty)
    )
:effect
    (and
        (cards-moving)
        (cards-moving-south)
        (not (card-at ?cm ?x ?y ))
        (new-headtail-card ?cm)
        (next-moving-card ?cnext)
        (increase (total-cost) (move-card))
    )
)

(:action move-card-south
:parameters(?cm - card ?x - gridpos ?y - gridpos  ?cnext - card ?nexty - gridpos ?prevy - gridpos)
:precondition
    (and
        (cards-moving)
        (cards-moving-south)
        (not (robot-at ?cm))
        (next-moving-card ?cm)
        (card-at ?cm ?x ?y )
        (card-at ?cnext ?x ?nexty)
        (next ?prevy ?y)
        (next ?y ?nexty)
    )
:effect
    (and
        (cards-moving)
        (cards-moving-south)
        (not (card-at ?cm ?x ?y))
        (card-at ?cm ?x ?prevy)
        (not (next-moving-card ?cm))
        (next-moving-card ?cnext)
        (increase (total-cost) (move-card))
    )
)

(:action stop-move-card-south
:parameters(?cm - card ?x - gridpos ?y - gridpos ?prevy - gridpos ?newtc - card)
:precondition
    (and
        (cards-moving)
        (cards-moving-south)
        (not (robot-at ?cm))
        (next-moving-card ?cm)
        (card-at ?cm ?x ?y )
        (next ?prevy ?y)
        (min-pos ?y)
        (new-headtail-card ?newtc)
    )
:effect
    (and
        (not (cards-moving))
        (not (cards-moving-south))
        (not (card-at ?cm ?x ?y))
        (card-at ?cm ?x ?prevy)
        (card-at ?newtc ?x ?y)
        (not (new-headtail-card ?newtc))
        (not (next-moving-card ?cm))
        (increase (total-cost) (move-card))
    )
)

(:action leave
:parameters(?c - card ?prow - gridpos ?pcolumn - gridpos)
:precondition
    (and
        (not (cards-moving))
        (robot-at ?c)
        (card-at ?c ?prow ?pcolumn)
        (max-pos ?prow)
        (max-pos ?pcolumn)
        (not (blocked ?c s ))
    )
:effect
    (and
        (increase (total-cost) (move-card))
        (left)
    )
)
)

\end{lstlisting}

\subsection{Sokoban}

\begin{lstlisting}[caption=Sokoban domain file]
(define (domain typed-sokoban)
(:requirements :typing)
(:types LOC DIR)
(:predicates 
             (at-robot ?l - LOC)
             (has-box ?l - LOC)
             (adjacent ?l1 - LOC ?l2 - LOC ?d - DIR) 
             (clear ?l - LOC)
)


(:action move
:parameters (?from - LOC ?to - LOC ?dir - DIR)
:precondition (and (clear ?to) (at-robot ?from) (adjacent ?from ?to ?dir))
:effect (and (at-robot ?to) (not (at-robot ?from)))
)
             

(:action push
:parameters  (?rloc - LOC ?bloc - LOC ?floc - LOC ?dir - DIR)
:precondition (and (at-robot ?rloc) (has-box ?bloc) (clear ?floc)
	           (adjacent ?rloc ?bloc ?dir) (adjacent ?bloc ?floc ?dir))

:effect (and (at-robot ?bloc) (has-box ?floc) (clear ?bloc)
             (not (at-robot ?rloc)) (not (has-box ?bloc)) (not (clear ?floc)))
)
\end{lstlisting}

\end{document}